\documentclass{article} 
\usepackage{iclr2022_conference,times}

\usepackage{amsmath,amsfonts,bm}

\def\eqref#1{equation~\ref{#1}}

\def\1{\bm{1}}

\DeclareMathAlphabet{\mathsfit}{\encodingdefault}{\sfdefault}{m}{sl}
\SetMathAlphabet{\mathsfit}{bold}{\encodingdefault}{\sfdefault}{bx}{n}

\DeclareMathOperator*{\argmin}{arg\,min}

\usepackage{hyperref}
\usepackage{url}
\usepackage{graphicx}

\title{Finding Biological Plausibility for Adversarially Robust Features via Metameric Tasks}

\author{Anne Harrington \& Arturo Deza\\
Center for Brains, Minds and Machines\\
Massachusetts Institute of Technology\\
\texttt{\{annekh,deza\}@mit.edu}\\
}

\iclrfinalcopy 
\begin{document}

\maketitle

\begin{abstract}
Recent work suggests that feature constraints in the training datasets of deep neural networks (DNNs) drive robustness to adversarial noise \citep{ilyas2019adversarial}.  The representations learned by such adversarially robust networks have also been shown to be more human perceptually-aligned than non-robust networks via image manipulations \citep{santurkar2019image,engstrom2019adversarial}.  Despite appearing closer to human visual perception, it is unclear if the constraints in robust DNN representations match biological constraints found in human vision.  Human vision seems to rely on texture-based/summary statistic representations in the periphery, which have been shown to explain phenomena such as crowding \citep{balas2009summary} and performance on visual search tasks \citep{rosenholtz2012summary}. To understand how adversarially robust optimizations/representations compare to human vision, we performed a psychophysics experiment using a metamer task similar to \citet{freeman2011metamers,wallis2019image,deza2019towards} where we evaluated how well human observers could distinguish between images synthesized to match adversarially robust representations compared to non-robust representations and a texture synthesis model of peripheral vision (Texforms \citep{long2018mid}).  We found that the discriminability of robust representation and texture model images decreased to near chance performance as stimuli were presented farther in the periphery.  Moreover, performance on robust and texture-model images showed similar trends within participants, while  performance on non-robust representations changed minimally across the visual field.  These results together suggest that (1) adversarially robust representations capture peripheral computation better than non-robust representations and (2) robust representations capture peripheral computation similar to current state-of-the-art texture peripheral vision models. More broadly, our findings support the idea that localized texture summary statistic representations may drive human invariance to adversarial perturbations and that the incorporation of such representations in DNNs could give rise to useful properties like adversarial robustness. Link to Code/Data:  \href{https://github.com/anneharrington/Adversarially-Robust-Periphery}{https://github.com/anneharrington/Adversarially-Robust-Periphery}.

\end{abstract}

\section{Introduction}


Texture-based summary statistic models of the human periphery have been shown to explain key phenomena such as crowding \citep{balas2009summary,freeman2011metamers} and performance on visual search tasks \citep{rosenholtz2012summary} when used to synthesize feature-matching images. These analysis-by-synthesis models have also been used to explain mid-level visual computation (\textit{e.g.} V2) via perceptual discrimination tasks on images for humans and primates~\citep{freeman2011metamers,ziemba2016selectivity,long2018mid}. 

However, while summary statistic models can succeed at explaining peripheral computation in humans, they fail to explain foveal computation and core object recognition that involve other representational strategies~\citep{logothetis1995shape,riesenhuber1999hierarchical,dicarlo2007untangling,hinton2021represent}. Modelling foveal vision with deep learning indeed has been the focus of nearly all object recognition systems in computer vision (as machines do not have a periphery)~\citep{lecun2015deep,schmidhuber2015deep} -- yet despite their overarching success in a plethora of tasks, they are  vulnerable to adversarial perturbations. This phenomena indicates: 1) a critical failure of current artificial systems~\citep{goodfellow2014explaining,szegedy2013intriguing}; and 2) a perceptual mis-alignment of such systems with humans~\citep{golan2019controversial,feather2019metamers,firestone2020performance,geirhos2021partial,funke2021five} -- with some exceptions~\citep{elsayed2018adversarial}. Indeed, there are many strategies to alleviate these sensitivities to perturbations, such as data-augmentation~\citep{rebuffi2021fixing,gowal2021improving}, biologically-plausible inductive biases~\citep{dapello2020simulating,reddy2020biologically,jonnalagadda2021foveater}, and adversarial training~\citep{tsipras2018robustness,madry2017towards}. This last strategy in particular (adversarial training) is popular, but has been criticized as being non-biologically plausible -- despite yielding some perceptually aligned images when inverting their representations~\citep{engstrom2019adversarial,santurkar2019image}.


\begin{figure}
    \centering
    \includegraphics[scale=1]{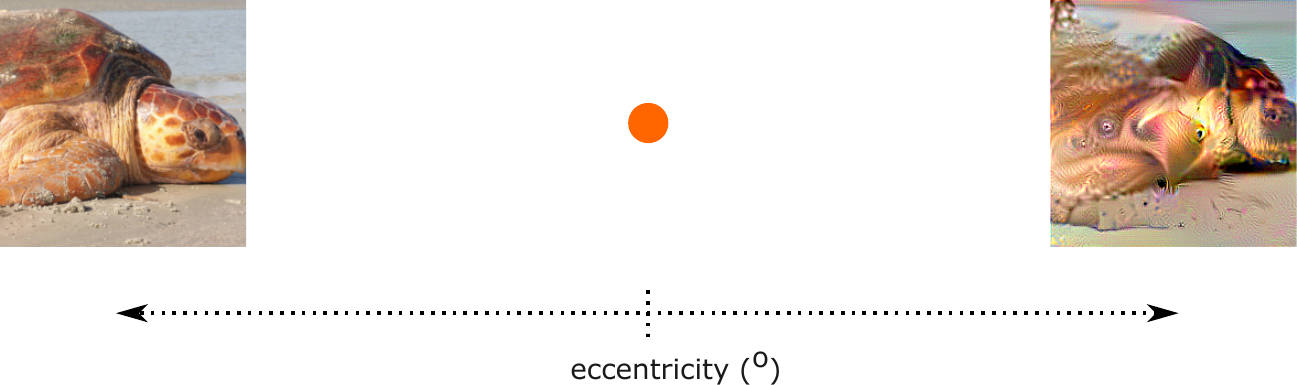}
    \caption{A sample un-perturbed (left) and synthesized adversarially robust (right) image are shown peripherally. When a human observer fixates at the orange dot (center), both images -- now placed away from the fovea -- are perceptually indistinguishable to each other (i.e. \textit{metameric}). In this paper we investigate if there is a relationship between peripheral representations in humans and learned representations of adversarially trained networks in machines in an analysis-by-synthesis approach. We psychophysically test this phenomena over a variety of images synthesized from an adversarially trained network, a non-adversarially trained network, and a model of peripheral computation as we manipulate retinal eccentricity over 12 humans subjects.
    }
    \label{fig:turtles_in_periphery}
\end{figure}

Motivated by prior work on summary statistic models of peripheral computation and their potential resemblance to inverted representations of adversarially trained networks, we wondered if these two apparently disconnected phenomena from different fields share any similarities (See Figure~\ref{fig:turtles_in_periphery}). Could it be that adversarially trained networks are robust because they encode object representations similar to human peripheral computation? We know machines do not have peripheral computation~\citep{azulay2019deep,deza2020emergent,alsallakh2021mind}, yet are susceptible to a type of adversarial attacks that humans are not. We hypothesize that object representation arising in human peripheral computation holds a critical role for high level robust vision in perceptual systems, but testing this has not been done.

Thus, the challenge we now face is how to compare an adversarially trained neural network model to current models of peripheral/mid-level visual processing -- and ultimately to human observers as the objective ground-truth. However, determining such perceptual parameterizations is computationally intractable. Inspired by recent works that test have tested summary statistic models via metameric discrimination tasks~\citep{deza2019towards,wallis2016testing,wallis2017parametric,wallis2019image}, we can evaluate how well the adversarially robust CNN model approximates the types of computations present in human peripheral vision with a set of rigorous psychophysical experiments wrt synthesized stimuli. 

Our solution consists of performing a set of experiments where we will evaluate the rates of human perceptual discriminability as a function of retinal eccentricity across the synthesized stimuli from an adversarially trained network vs synthesized stimuli from models of mid-level/peripheral computation. If the decay rates at which the perceptual discriminability across different stimuli are similar, then this would suggest that the transformations learned in an adversarially trained network are related to the transformations done by models of peripheral computation -- and thus, to the human visual system. It is worth noting that although adversarially robust representations have been shown to be more human-perceptually aligned \citep{ilyas2019adversarial,engstrom2019adversarial,santurkar2019image}, they still look quite different when placed in the foveal region from the original reference image~\citep{feather2021adversarial}. However, our eccentricity-varying psychophysical experiments are motivated by empirical work that suggests that the human visual periphery represents input in a texture-like scrambled way, that can appear quite different than how information is processed in the fovea~\citep{rosenholtz2016capabilities,stewart2020review,herrera2021flexible}.

\section{Synthesizing Stimuli as a window to Model Representation}

\begin{figure}[!t]
    \centering
    \includegraphics[scale=0.6]{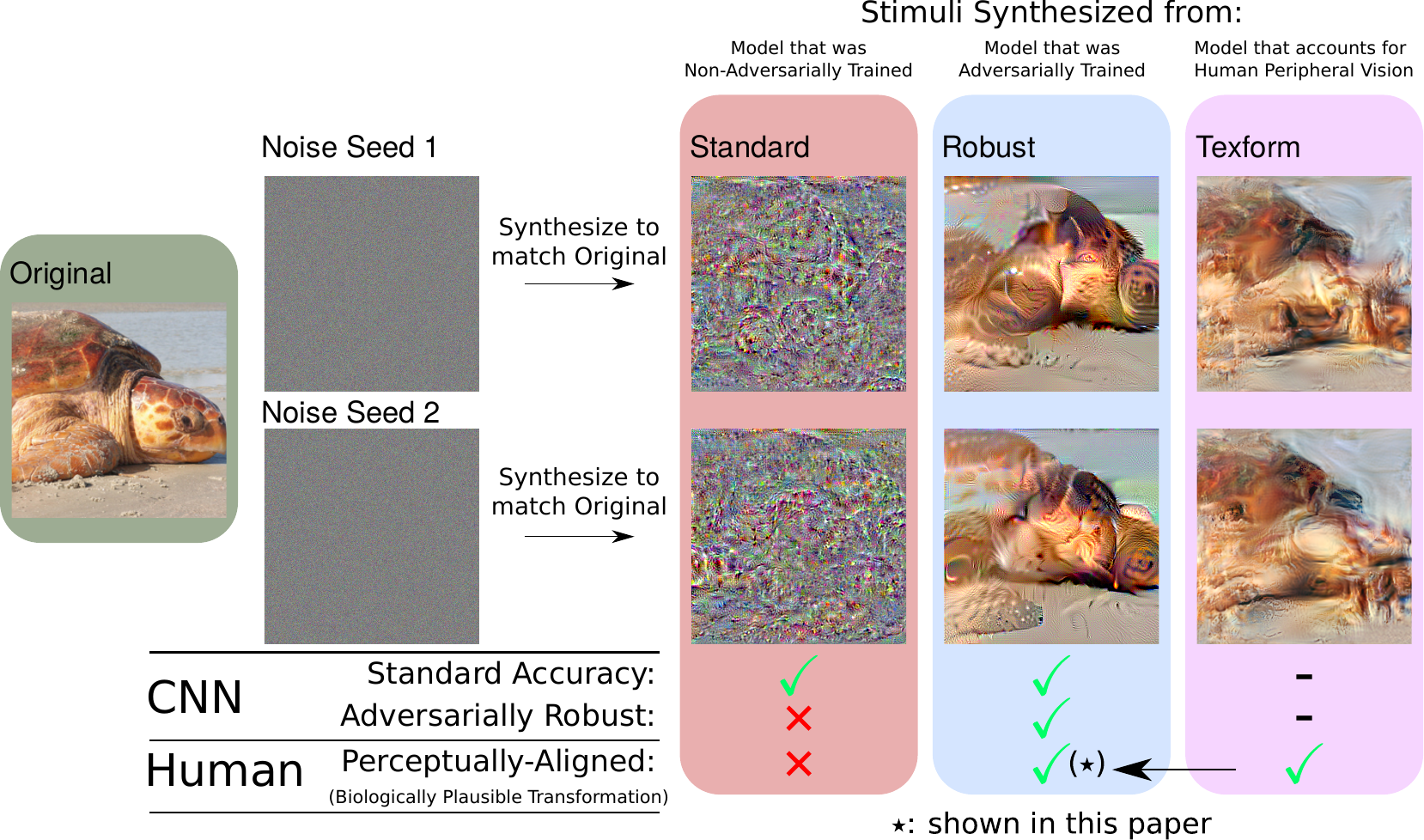}
    \caption{A sub-collection of different synthesized stimuli used in our experiments that shows the differences across (columns) and within (rows) perceptual models. The original stimuli is shown on the left, while two parallel Noise Seeds, give rise to different samples for the Standard, Robust and Texform stimuli. Critically, an adversarially trained network -- which was used to synthesize the Robust stimuli~\citep{engstrom2019adversarial} -- has implicitly learned to encode a structural prior with localized texture-like distortions similar to the physiologically motivated Texforms that account for  several phenomena of \textit{human peripheral computation}~\citep{freeman2011metamers,rosenholtz2012summary,long2018mid}. However, Standard stimuli, which are images synthesized from a network with Regular (Non-Adversarial) training have no resemblance to the original sample. In this paper we evaluate how similar these models are, via their derived stimuli, with a set of controlled human psychophysics experiments where we vary the retinal eccentricity of the stimuli.}
    \label{fig:Stimuli_Collection}
\end{figure}

Suppose we have the functions $g_\text{Adv}(\circ)$ and $g_\text{Standard}(\circ)$ that represent the adversarially trained and standard (non-adversarially) trained neural networks; how can we compare them to human peripheral computation if the function $g_\text{Human}(\circ)$ is computationally intractable?

One solution is to take an analysis-by-synthesis approach and to synthesize a collection of stimuli that match the feature response of the model we'd like to analyze -- this is also known as feature inversion~\citep{mahendran2015understanding,feather2019metamers}. If the inverted features (stimuli) of two models are perceptually similar, then it is likely that the learned representations are also aligned. For example, if we'd like to know what is the stimuli $x'$ that produces the same response to the stimuli $x$ for a network $g'(\circ)$, we can perform the following minimization:
\begin{equation}
\label{eq:function_g}
x' = \argmin_{x_0}[||g'(x)-g'(x_0)||_2]
\end{equation}
In doing so, we find $x'$ which should be different from $x$ for a non-trivial solution. This is known as a metameric constraint for the stimuli pair $\{x,x_0\}$ wrt to the model $g'(\circ)$ : $g'(x)=g'(x')$ s.t. $x\neq x'$ for a starting pre-image $x_0$ that is usually white noise in the iterative minimization of Eq.\ref{eq:function_g}. Indeed, for the adversarially trained network of~\cite{ilyas2019adversarial,engstrom2019adversarial,santurkar2019image}, we can synthesize robust stimuli wrt to the original image $x$ via:
\begin{equation}
\label{eq:function_g_Adv}
\tilde{x} = \argmin_{x_0}[||g_\text{Adv}(x)-g_\text{Adv}(x_0)||_2]
\end{equation}
which implies -- if the minimization goes to zero -- that:
\begin{equation}
\label{eq:Adv_1}
||g_\text{Adv}(x)-g_\text{Adv}(\tilde{x})||_2=0
\end{equation}
Recalling the goal of this paper, we'd like to investigate if the following statement is true: \textit{``a transformation resembling peripheral computation in the human visual system can closely be approximated by an adversarially trained network''}, which is formally translated as: $g_\text{Adv}\sim g_\text{Human}^{r_*}$ for some retinal eccentricity $(r_*)$, then from Eq.~\ref{eq:Adv_1} we can also derive:
\begin{equation}
\label{eq:Human_1}
||g_\text{Human}^{r_*}(x)-g_\text{Human}^{r_*}(\tilde{x})||_2=0
\end{equation}
However, $g_\text{Human}(\circ)$ is computationally intractable, so how can we compute Eq.\ref{eq:Human_1}? A first step is to perform a psychophysical experiment such that we find a retinal eccentricity $r_*$ at which human observers can not distinguish between the original and synthesized stimuli -- thus behaviourally proving that the condition above holds, without the need to directly compute $g_\text{Human}$.

More generally, we'd like to compare the \textit{psychometric functions} between stimuli generated from a standard trained network (standard stimuli), an adversarially trained network (robust stimuli), and a model that captures peripheral and mid-level visual computation (texform stimuli~\citep{freeman2011metamers,long2018mid}). Then we will assess how the psychometric functions vary as a function of retinal eccentricity. If there is significant overlap between psychometric functions between one model wrt the model of peripheral computation; then this would suggest that the transformations developed by such model are similar to those of human peripheral computation. We predict that this will be the case for the adversarially trained network $(g_\text{Adv}(\circ))$. Formally, for any model $g$, and its synthesized stimuli $x_g$ -- as shown in Figure~\ref{fig:Stimuli_Collection}, we will define the psychometric function $\delta_\text{Human}$, which depends on the eccentricity $r$ as:
\begin{equation}
\delta_\text{Human}(g;r)=||g_\text{Human}^{r}(x)-g_\text{Human}^{r}(x_g)||_2
\end{equation}
where we hope to find:
\begin{equation}
\label{eq:Psychometric_Equals}
\delta_\text{Human}(g_\text{Adv};r) = \delta_\text{Human}(g_\text{Texform};r) ; \forall r. 
\end{equation}

\subsection{Standard and Robust Model Stimuli}
To evaluate robust vs non-robust feature representations, we used the ResNet-50 models of \citet{santurkar2019image,ilyas2019adversarial, engstrom2019adversarial}.  We used their models so that our results could be interpreted in the context of their findings that features may drive robustness.   Both models were trained on a subset of ImageNet~\citep{russakovsky2015imagenet}, termed Restricted ImageNet (Table \ref{data_classes}).  The benefit of Restricted ImageNet, stated by \citeauthor{ilyas2019adversarial,engstrom2019adversarial}, is models can achieve better standard accuracy than on all of ImageNet.  One drawback is that it is imbalanced across classes.  Although the class imbalance was not problematic for comparing the adversarially robust model to standard-trained one, we did ensure that there was a nearly equal number of images per class when selecting images for our stimulus set to avoid class effects in our experiment (i.e. people are better at discriminating dog examples than fishes independent of the model training).

Using their readily available models, we synthesized robust and standard model stimuli using an image inversion procedure \citep{mahendran2015understanding,gatys2015texture,santurkar2019image,engstrom2019adversarial,ilyas2019adversarial}.  We used gradient descent to minimize the difference between the representation of the second-to-last network layer of a target image and an initial noise seed as shown in Figure~\ref{fig:synthesis_prodecure}. Target images were randomly chosen from the test set of Restricted ImageNet. We chose $100$ target images for each of the $9$ classes and synthesized a robust and standard stimulus for $2$ different noise seeds.  $5$ target images were later removed as they were gray-scale and could not also be rendered as Texforms with the same procedure as the majority.  All stimuli were synthesized at a size of 256 $\times$ 256 pixels, this was equivalent to $6.67\times6.67$ degrees of visual angle (d.v.a.) when performing the psychophysical experiments (See~\ref{sec:Apparatus} for calculation).

\subsection{Texform Stimuli}

\begin{figure}[!t]
    \centering
    \includegraphics[width=1.0\columnwidth]{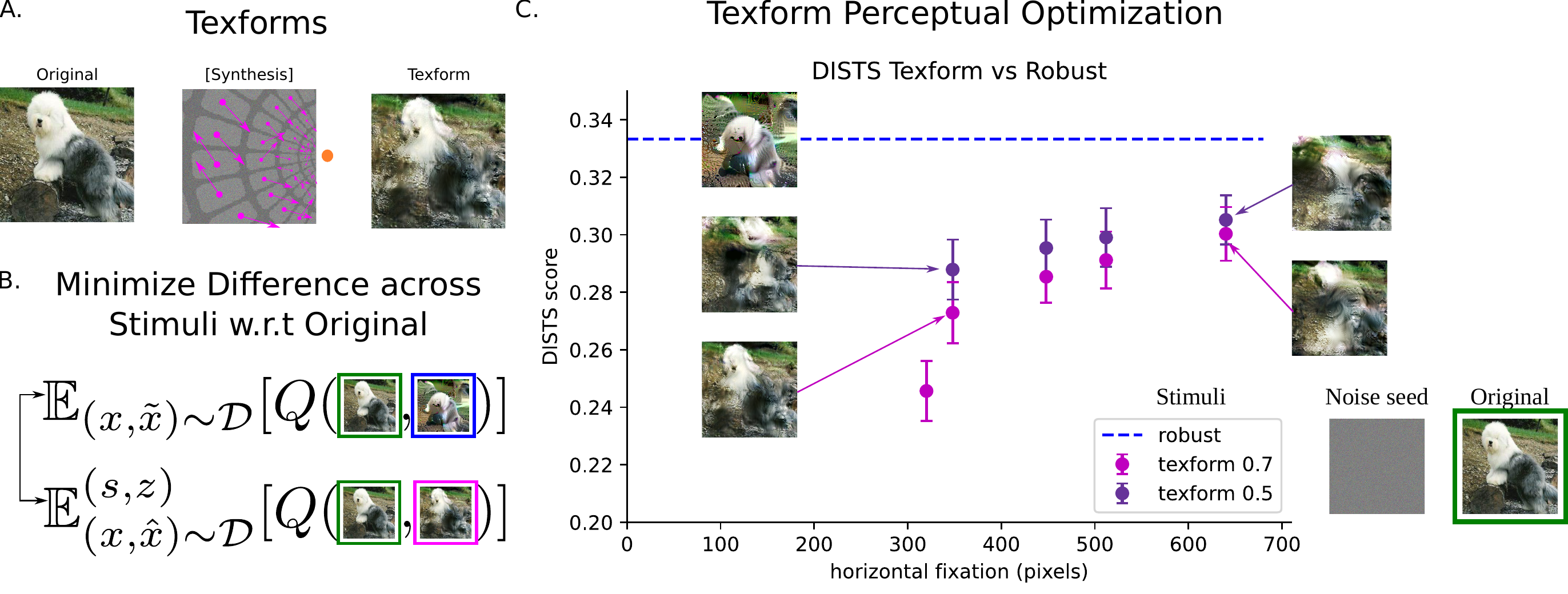}
    \caption{(A.) A cartoon depicting the texform generating process where log-polar receptive fields are used as areas over which localized texture synthesis is performed -- imitating the type of texture-based computation found in the human periphery and area V2. (B.) The perceptual optimization framework where the goal is to find the set of texform parameters $(s_*,z_*)$ over which the loss is minimized to match the levels of distortions of the robust stimuli \textit{before} performing human psychophysics. (C.) The texform perceptual optimization pipeline results show the DISTS scores~\citep{ding2020image} of texforms synthesized across different scaling factors and fixations points compared to adversarially robust stimuli synthesized from the same noise seed across 45 images (5 per RestrictedImageNet class selected randomly). Error bars indicate two standard errors from the mean.}
    \label{fig:texform_perceptual_opt}
\end{figure}

Texforms~\citep{long2018mid} are object-equivalent rendered stimuli from the~\cite{freeman2011metamers,rosenholtz2012summary} models that break the metameric constraint to test for mid-level visual representations in Humans. These stimuli -- initially inspired by the experiments of~\cite{balas2009summary} -- preserve the coarse global structure of the image and its localized texture statistics~\citep{portilla2000parametric}. Critically, we use the texform stimuli -- \textit{voiding the metameric constraint} -- as a perceptual control for the robust stimuli, as the texforms incarnate a sub-class of biologically-plausible distortions that loosely resemble the mechanisms of human peripheral processing.

As the texform model has 2 main parameters which are the scaling factor $s$ and the simulated point of fixation $z$, we must perform a perceptual optimization procedure to find the set of texforms $\hat{x}$ that match the robust stimuli $\tilde{x}$ as close as possible (w.r.t to the original image) \textit{before} testing their discriminability to human observers as a function of eccentricity. To do this, we used the accelerated texform implementation of~\cite{deza2019accelerated} and generated 45 texforms with the \textit{same} collection of initial noise seeds as the robust stimuli to be used as perceptual controls. Similar to~\cite{deza2020emergent} we minimize the perceptual dissimilarity $\mathcal{Z}$ to find $(s_*,z_*)$ over this subset of images that we will later use in the human psychophysics ($\sim$ 900 texforms):

\begin{equation}
\label{eq:Texform_Optimization}
    (s_*,z_*)=\argmin_{(s,z)}\mathcal{Z} = || \mathbb{E}_{(x,\tilde{x})\sim\mathcal{D}}[Q(x,\tilde{x})]-\mathbb{E}_{(x,\hat{x})\sim\mathcal{D}}^{(s,z)}[Q(x,\hat{x})] ||_2
\end{equation}

for an image quality assessment (IQA) function $Q(\circ,\circ)$. We selected DISTS in our perceptual optimization setup given that it is the IQA metric that is most tolerant to texture-based transformations~\citep{ding2020image,ding2021comparison}. A cartoon illustrating the texform rendering procedure, the perceptual optimization framework and the respective results can be seen in Figure~\ref{fig:texform_perceptual_opt}. In our final experiments (See Next Section) we used texforms rendered with a simulated scale of 0.5 and horizontal simulated point of fixation placed at 640 pixels. Critically, this value is \textit{immutable} and texforms (like robust stimuli) will not vary as a function of eccentricity to provide a fair discriminability control in the human psychophysics. For a further discussion on texforms and their biological plausibility and/or synthesis procedure, please see Supplement~\ref{sec:Sup_Texforms}.

\section{Human Psychophysics: Discriminating between stimuli as a function of retinal eccentricity}
\label{sec:Stimuli}

\begin{figure}[!t]
    \centering
    \includegraphics[width=\columnwidth]{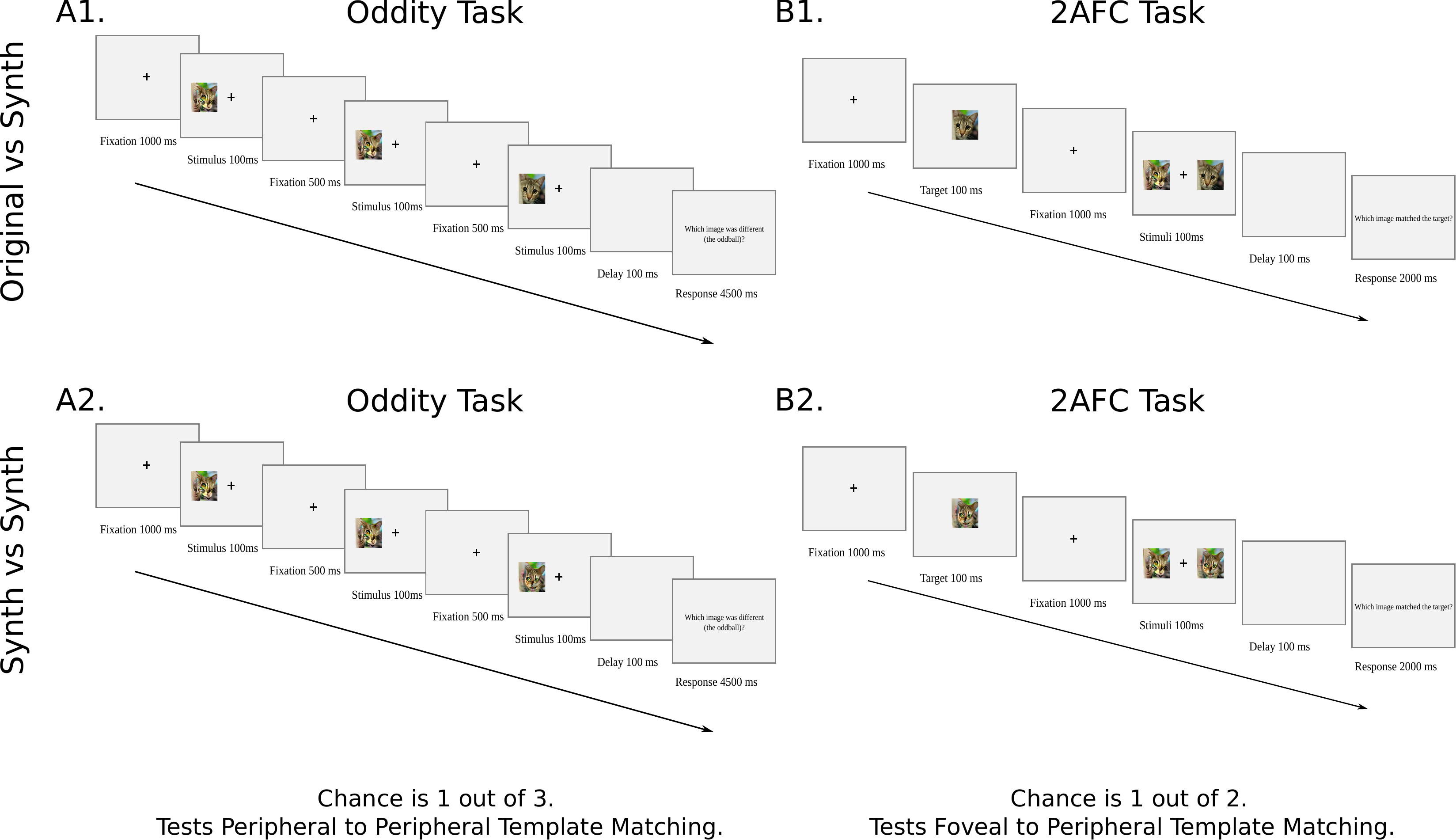}
    \caption{A schematic of the two human psychophysics experiments conducted in our paper. The first (A1.,A2.) illustrates an Oddity task where observers must determine the `oddball' stimuli without moving their eyes for very brief presentation times (100 ms) that are masked which do not allow for eye-movements or feedback processing. The second experiment (B1.,B2.) shows the 2 Alternative Forced Choice (AFC) Matching Tasks where observers must match the foveal template to 2 potential candidates on the left or right of the image. All trials are done while observers are instructed to remain fixating at the center of the image. These two experiments test different mechanisms of visual processing. In particular the Oddity task examines discriminability of purely peripheral-to-peripheral representations, while the 2AFC task evaluates discriminability of foveal-to-peripheral representation. Differences across rows indicate the type of interleaved trials shown to the observers: (1) Original vs Synthesized, and (2) Synthesized vs Synthesized. Critically these are image perceptual discrimination tasks \textit{not} image categorization tasks.}
    \label{fig:psychophysics_tasks}
\end{figure}

We designed two human psychophysical experiments: the first was a an oddity task similar to \citet{wallis2016testing}, and the second was a matching, two-alternative forced choice task (2AFC).  Two different tasks were used to evaluate how subjects viewed synthesized images both only in the periphery (oddity) and those they saw in the fovea (matching 2AFC). The oddity task consisted of finding the oddball stimuli out of a series of 3 stimuli shown peripherally one after the other (100ms) masked by empty intervals (500ms) while holding center fixation. Chance for the oddity task was 1 out of 3 $(33.3\%)$. The matching 2AFC task consisted of viewing a stimulus in the fovea (100ms) and then matching it to two candidate templates in the visual periphery (100 ms) while holding fixation. A 1000 ms mask was used in this experiment and chance was 50\%. 

For both experiments, we also had interleaved trials where observers had to engage in an Original stimuli vs Synthesized stimuli task, or a Synthesized stimuli vs Synthesized stimuli discrimination task (two stimulus pairs synthesized from \textit{different} noise seeds to match model representations). The goal of these experimental variations (called \textit{`stimulus roving'}) was two-fold: 1) to add difficulty to the tasks thus reducing the likelihood of ceiling effects; 2) to gather two psychometric functions per family of stimuli, which portrays a better description of each stimuli's evoked perceptual signatures.

We had 12 participants complete both the oddity and matching 2AFC experiments.  The oddity task was always performed first so that subjects would never have foveated on the images before seeing them in the periphery.  We had two stimulus conditions (1) robust \& standard model images and (2) texforms.  Condition 1 consisted of the inverted representations of the adversarially robust and standard-trained models.  The two model representations were randomly interleaved since they were synthesized with the same procedure.  Condition 2 consisted of texforms synthesized with a fixed and perceptually optimized fixation and scaling factor which yielded images closest in structure to the robust representations at foveal viewing (robust features have no known fixation and scaling -- which is why partly we evaluate multiple points in the periphery. Recall Figure~\ref{fig:texform_perceptual_opt}).  We randomly assigned the order in which participants saw the different stimuli. More details found in~\ref{sec:Apparatus}.  

\begin{figure}[!t]
    \centering
    \includegraphics[width=1.0\columnwidth]{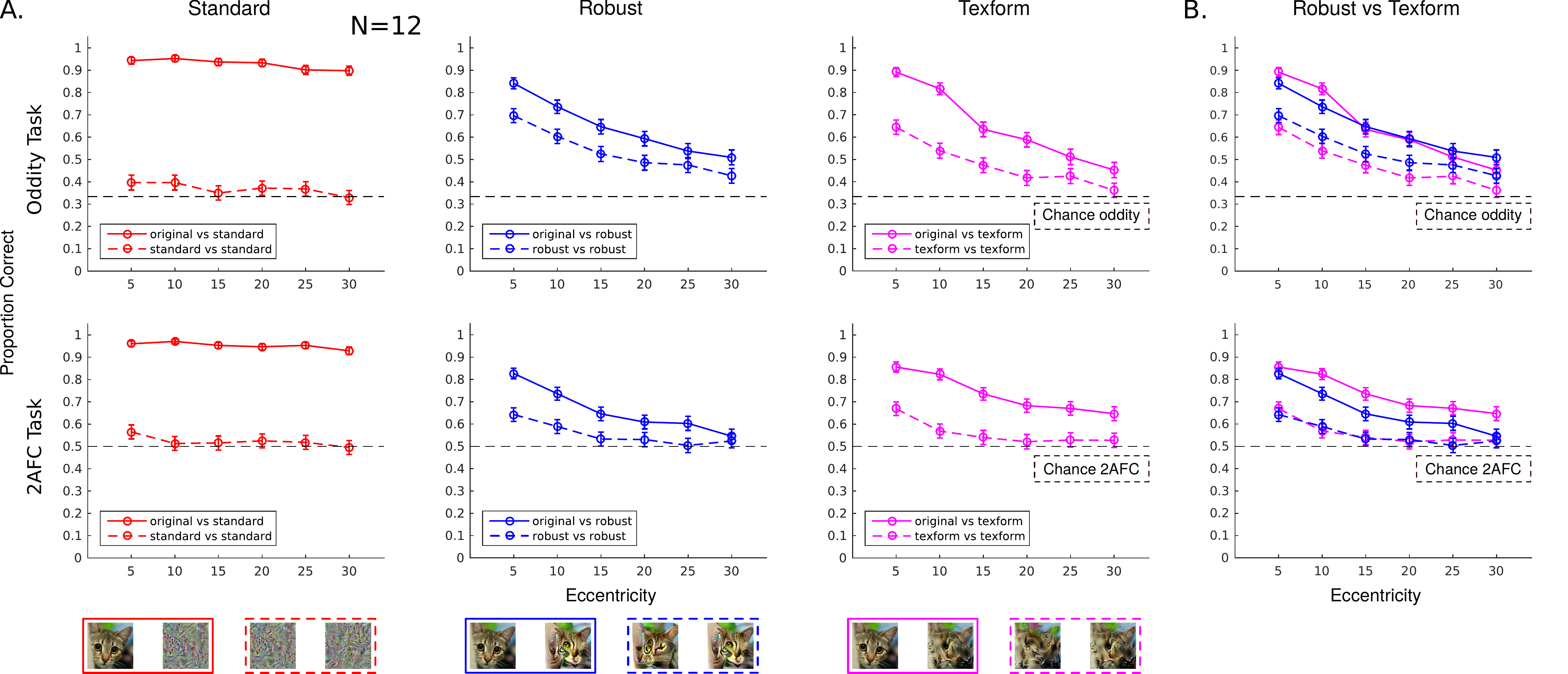}
    \caption{Pooled observer results of both psychophysical experiments are shown (top and bottom row). (A.) Left: we see that observers perfectly discriminate the original image wrt the standard stimuli, in addition to chance performance when comparing against synthesized stimuli. Critically there is no interaction of the standard stimuli with retinal eccentricity which suggests that the model used to synthesize such stimuli is a poor model of peripheral computation. Middle: Human observers do worse at discriminating the robust stimuli wrt the original as a function of eccentricity and also between synthesized robust samples. Given this decay in perceptual discriminability, it would suggest that the adversarially trained model used to synthesize robust stimuli does capture aspects of peripheral computation. This effect can also be seen on the texforms (Right) -- which have been extensively used as stimuli from derived models that capture peripheral and V2-like computation. (B.) Superimposed human performance for Robust and Texform stimuli. Errorbars are computed via bootstrapping and represent the 95\% confidence interval.}
    \label{fig:summary_fig_1}
\end{figure}

The main results of our 2 experiments can be found in Figure~\ref{fig:summary_fig_1}, where we show how well Humans can discriminate per type of stimuli class and task. Mainly, human observers achieve near perfect discrimination rates for the Standard stimuli wrt to their original references, but near chance levels when discriminating to another synthesized sample. This occurs for both experimental paradigms (Oddity + 2AFC), suggesting that the network responsible for encoding standard stimuli is a poor model of human peripheral vision given no interaction with retinal eccentricity.

However, we observe that Humans show similar perceptual discriminability rates for Robust and Texform stimuli -- and that these vary in a similar way as a function of retinal eccentricity. Indeed, for both of these stimuli their perceptual discrimination rates follow a sigmoidal decay-like curve when comparing the stimuli to the original, and also between synthesized samples. The similarity between the blue and magenta curves from Figure~\ref{fig:summary_fig_1} suggests that if the texform stimuli do capture some aspect of peripheral computation, then -- by transitivity -- so do the adversarial stimuli which were rendered from an adversarially trained network. These results empirically verify our initial hypothesis that we set out to test in this paper. A superposition of these results in reference to the Robust stimuli for a better interpretation can also be seen in Figure~\ref{fig:summary_fig_1} (B.). 

\vspace{-10pt}
\subsection{Simulated Fovea/Periphery Image Quality Assessment (IQA) across stimuli}
\label{sec:Computational}
Some distortions are more perceptually noticeable than others for humans and neural networks \citep{berardino2017eigen,martinez2019praise} -- so how do we assess which model better accounts for peripheral computation, if there are many distortions (derived from the synthesized model stimuli) that can potentially yield the same perceptual sensitivity in a discrimination task? 

Our approach consists of computing two IQA metrics (DISTS \& MSE) over the entire psychophysical testing set over 2 opposite levels of a Gaussian Pyramid decomposition~\citep{burt1987laplacian}. This procedure checks which stimuli presents the greatest distortion (MSE), and yet yields greater perceptual invariance (DISTS). A Gaussian Pyramid decomposition was selected as it stimulates the frequencies preserved given changes in human contrast sensitivity and cortical magnification factor from fovea to periphery~\citep{anstis1974chart,geisler1998real}. These two metrics were one that is texture-tolerant and perceptually aligned (DISTS), and another that is a non-perceptually aligned metric: Mean Square Error (MSE). Both IQA metrics were computed in pixel space for both the Original vs Synthesized and Synthesized vs Synthesized conditions.

\begin{figure}[!t]
    \centering
    \includegraphics[width=0.9\columnwidth]{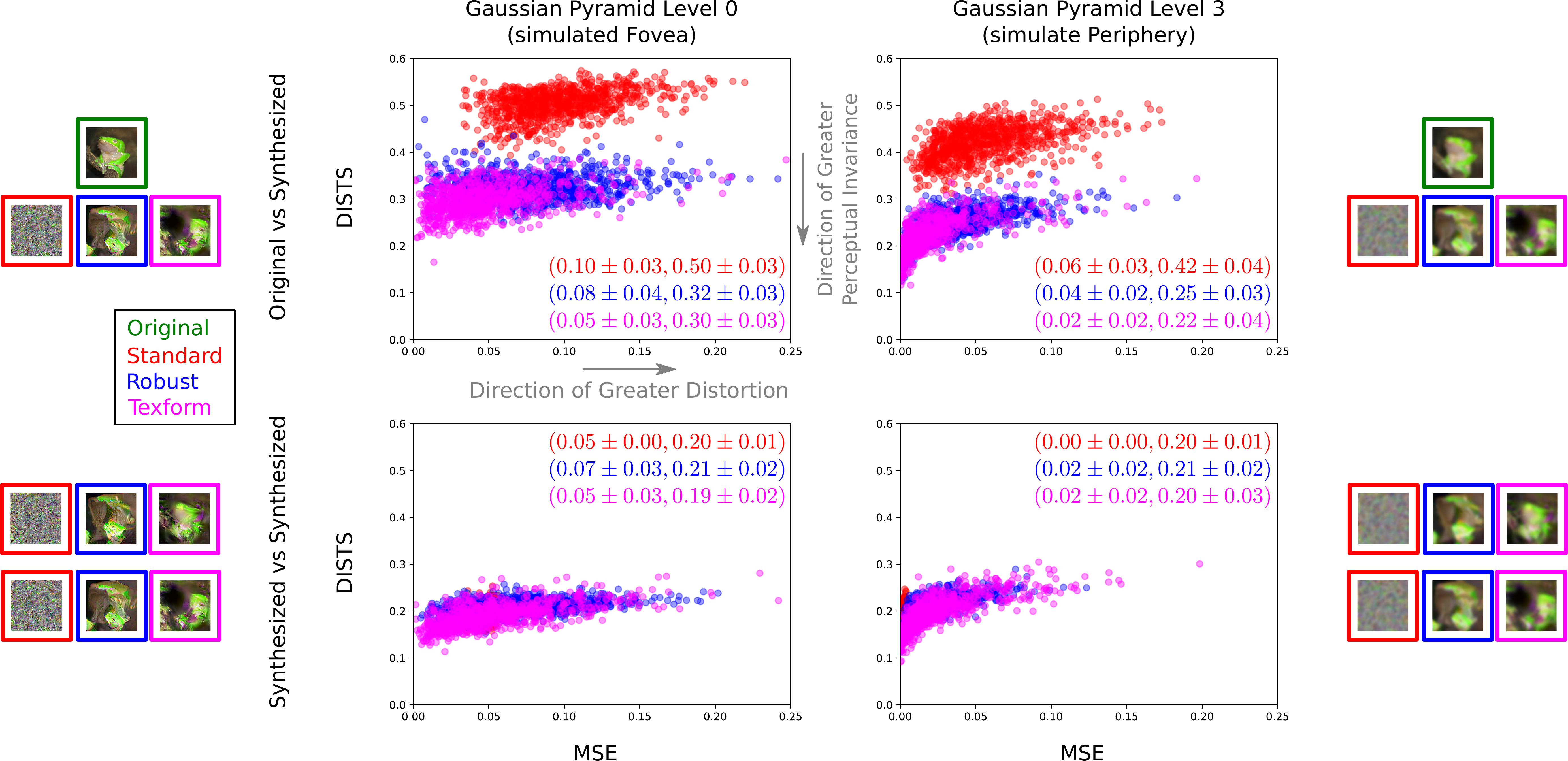}
    \vspace{-5pt}
    \caption{Here we evaluate how the different stimuli differ to each other wrt to the original (top row) or synthesized samples (bottom row) via two IQA metrics: DISTS and MSE. This characterization allows us to compare which model discards more information (MSE) while yielding a greater degree of model based perceptual invariance. We find that Texform and Robust stimuli are similar terms of both IQA scores, suggesting their models compute the same transformations. This is observed at the 0th level (simulated fovea) and 3rd level (simulated periphery) of the Gaussian Pyramid.}
    \label{fig:Blur_Operator}
\end{figure}

Results are explained in Figure~\ref{fig:Blur_Operator}, where Standard Stimuli yields low perceptual invariance to the original image at both levels of the Gaussian Pyramid, but robust and texform stimuli have a similar degree of perceptual invariance. Critically, robust stimuli are slightly more distorted via MSE than texform stimuli suggesting that the adversarially trained model has learned to represent peripheral computation better than the texform model by maximizing the perceptual null space and throwing away more useless low-level image features (hence achieving greater Mean Square Error).

\begin{figure}[!t]
    \centering
    \includegraphics[width=0.9\columnwidth]{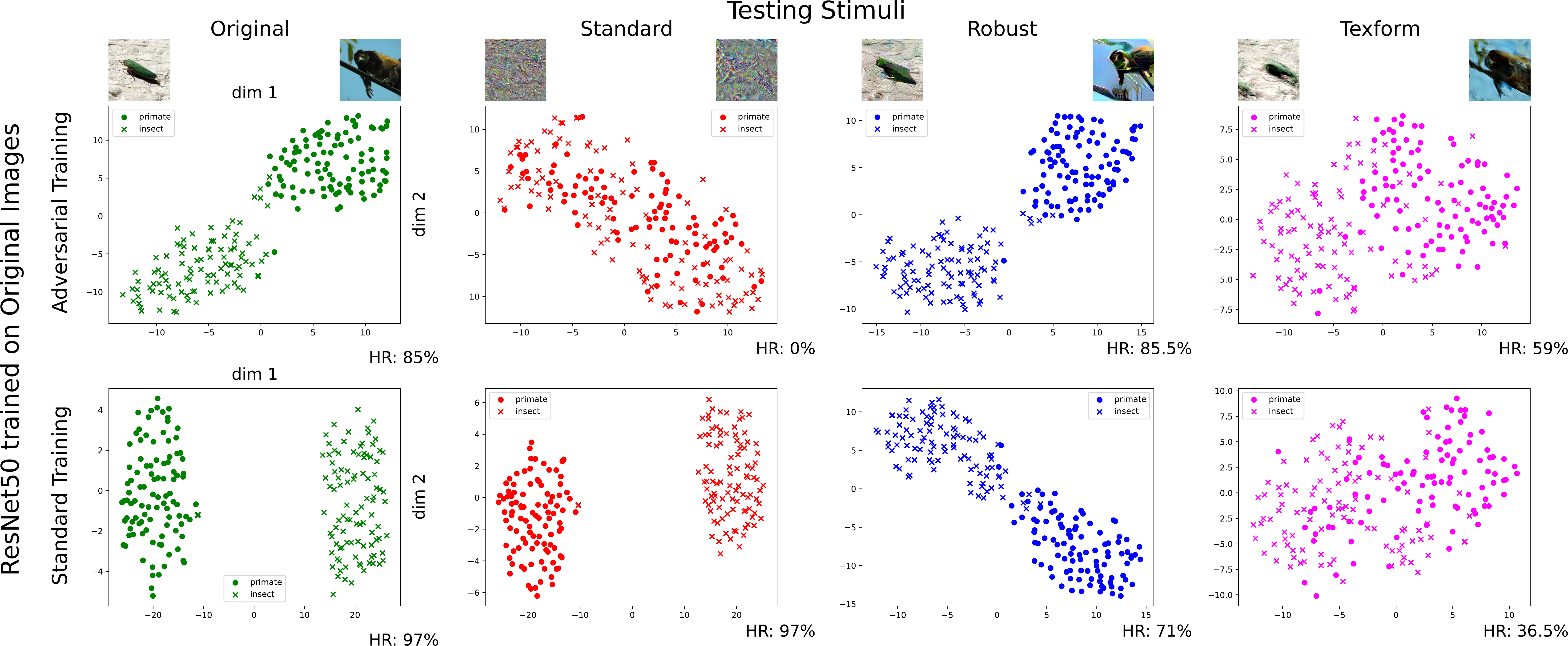}
    \vspace{-5pt}
    \caption{Here we show a 2D projection using t-SNE~\citep{van2008visualizing} to visualize the outputs of the last layer of the Adversarially trained network (that was used to synthesize the Robust Stimuli), and the Standard trained network (that was used to synthesize the Standard stimuli), both on a family of different stimuli: Original, Standard, Robust and Texform. The Adversarially trained network -- similar to the human -- can not distinguish between 2-class Standard Stimuli (unlike the Standard Network that has a near perfect 2-class hit rate). Most importantly, the Adversarially trained network yields a near double hit rate on Texform classification wrt the Standard trained network. This suggests that the Adversarially trained network has a representation that is more perceptually aligned to models of Peripheral Computation than the Standard trained model.}
    \label{fig:Projection}
\end{figure}
\vspace{-10pt}
\section{Discussion}
\label{sec:Discussion}

\begin{figure}[!t]
    \centering
    \includegraphics[width=0.9\columnwidth]{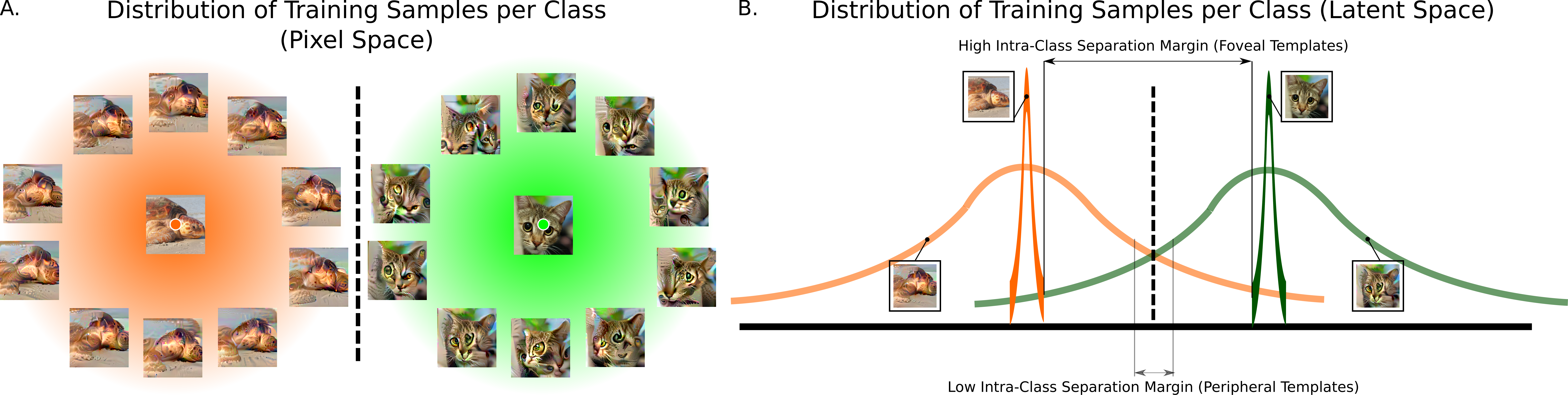}
    \caption{A cartoon depicting a conjecture of how peripheral computation may induce adversarial robustness. In (A.) we see a family of perceptually equidistant peripheral templates from the original foveal template (center dot) constructed with the adversarially robust model used to perform peripheral encoding. In (B.) we observe the same templates projected from a high-dimensional space into a uni-dimensional space. We also see that the greater \textit{covariances} only induced by peripheral templates lead to greater adversarial robustness during learning in a perceptual system -- despite having equal Intra-Class \textit{means} (for both foveal or peripheral templates). This suggests that peripheral computation may implicitly act as a \textit{natural visual regularizer} of learned representations.}
    \vspace{-10pt}
    \label{fig:turtles_and_cats}
\end{figure}

\vspace{-10pt}
One of the fundamental questions in modern computer vision is understanding what are the principles that give rise to adversarial stimuli -- which show a striking perceptual divergence between man and machine~\citep{feather2019metamers,golan2019controversial,geirhos2021partial,funke2021five}. While it may seem  theoretically impossible to escape from adversarial stimuli~\citep{gilmer2018adversarial} -- perhaps our efforts in the community should focus on understanding the biologically plausible mechanisms \textit{(if any)} of the solutions that grant some level of adversarial robustness aligned with human error. In this paper we focused on potentially linking the representations learned from an adversarially trained network and human peripheral computation via a series of psychophysical experiments through a family of stimuli synthesized from these models. 

We found that stimuli synthesized from an adversarially trained (and thus robust) network are metameric to the original stimuli in the further periphery (slightly above $30\deg$) for both Oddity and 2AFC Matching tasks. However, more important than deriving a critical eccentricity for metameric gaurantees across stimuli in Humans -- we found a surprisingly similar pattern of results in terms of how perceptual discrimination interacts with retinal eccentricity when comparing the adversarially trained network's robust stimuli with classical models of peripheral computation and V2 encoding (mid-level vision) that were used to render the texform stimuli~\citep{freeman2011metamers,long2018mid,ziemba2016selectivity,ziemba2021opposing}. Further, this type of eccentricity-driven interaction does not occur for stimuli derived from non-adversarially trained (standard) networks.

More generally, now that we found that adversarially trained networks encode a similar class of transformations that occur in the visual periphery -- how do we reconcile with the fact that adversarial training is biologically \textit{implausible} in humans? Recall from the work of~\cite{ilyas2019adversarial} that performing \textit{standard training} on robust images yielded similar generalization and adversarial robustness as performing adversarial training on standard images; how does this connect then to human learning if we assume a uniform learning rule in the fovea and the periphery? 

We think the answer lies in the fact that as humans learn to perform object recognition, they not only fixate at the target image, but they also look around, and can eventually learn where to make a saccade given candidate object peripheral templates -- thus learning certain invariances when the object is placed both in the fovea and the periphery~\citep{cox2005breaking,williams2008feedback,poggio2014computational,han2020scale}. This is an idea that dates back to~\cite{von1867handbuch}, as highlighted in~\cite{stewart2020review} on the interacting mechanisms of foveal and peripheral vision in humans.

Altogether, this could suggest that spatially-uniform high-resolution processing is redundant and sub-optimal in the \textit{o.o.d.} regime in the way that the visual representation which is computed is independent of point of fixation -- as seen classically in adversarially-vulnerable CNNs that are translation invariant and have no foveated/spatially-adaptive computation. Counter-intuitively, the fact that our visual system \textit{is} spatially-adaptive could give rise to a more robust encoding mechanism of the visual stimulus as observers can  encode a distribution rather than a point as they move their center of gaze~\citep{nandy2012saccade}. Naturally, from all the possible types of transformations, the ones that are similar to those shown in this paper -- which loosely resemble localized texture-computation -- are the ones that potentially lead to a robust hyper-plane during learning for the observer (See Fig.~\ref{fig:Projection} and~\ref{fig:turtles_and_cats}).

Finally, we'd like to add a disclaimer -- we use the term \textit{biological plausibility} at a representational level through-out this paper. However, current work is looking into reproducing the experiments carried out in this paper with a physiological component to explore temporal dynamics (MEG) and localization (fMRI) evoked from the stimuli. While it is not obvious if we will find a perceptual signature of the adversarial robust stimuli in humans, we think this novel stimuli and experimental paradigm presents a first step towards the road of linking what is known (and unknown) across texture representation, peripheral computation, and adversarial robustness in humans and machines.

\section*{Acknowledgements}
The authors would like to thank the Poggio, Rosenholtz, Simoncelli, DiCarlo \& Bonner labs and Lockheed Martin for valuable feedback. The authors would also like to thank Andrzej Banburski, Andrei Barbu, Tom Wallis, Pramod RT, Corey Ziemba and Tiago Marques for valuable discussions on the theory of distortions and suggestions for experimental controls. This work was sponsored by the Massachusetts Institute of Technology's Center for Brains, Minds \& Machines (MIT-CBMM), and Lockheed Martin Corporation.

\bibliography{iclr2022_conference}

\begin{thebibliography}{62}
\providecommand{\natexlab}[1]{#1}
\providecommand{\url}[1]{\texttt{#1}}
\expandafter\ifx\csname urlstyle\endcsname\relax
  \providecommand{\doi}[1]{doi: #1}\else
  \providecommand{\doi}{doi: \begingroup \urlstyle{rm}\Url}\fi

\bibitem[Alsallakh et~al.(2021)Alsallakh, Kokhlikyan, Miglani, Yuan, and
  Reblitz-Richardson]{alsallakh2021mind}
Bilal Alsallakh, Narine Kokhlikyan, Vivek Miglani, Jun Yuan, and Orion
  Reblitz-Richardson.
\newblock Mind the pad -- {\{}cnn{\}}s can develop blind spots.
\newblock In \emph{International Conference on Learning Representations}, 2021.
\newblock URL \url{https://openreview.net/forum?id=m1CD7tPubNy}.

\bibitem[Anstis(1974)]{anstis1974chart}
Stuart~M Anstis.
\newblock A chart demonstrating variations in acuity with retinal position.
\newblock \emph{Vision research}, 14\penalty0 (7):\penalty0 589--592, 1974.

\bibitem[Azulay \& Weiss(2019)Azulay and Weiss]{azulay2019deep}
Aharon Azulay and Yair Weiss.
\newblock Why do deep convolutional networks generalize so poorly to small
  image transformations?
\newblock \emph{Journal of Machine Learning Research}, 20:\penalty0 1--25,
  2019.

\bibitem[Balas et~al.(2009)Balas, Nakano, and Rosenholtz]{balas2009summary}
Benjamin Balas, Lisa Nakano, and Ruth Rosenholtz.
\newblock A summary-statistic representation in peripheral vision explains
  visual crowding.
\newblock \emph{Journal of vision}, 9\penalty0 (12):\penalty0 13--13, 2009.

\bibitem[Berardino et~al.(2017)Berardino, Laparra, Ball{\'e}, and
  Simoncelli]{berardino2017eigen}
Alexander Berardino, Valero Laparra, Johannes Ball{\'e}, and Eero Simoncelli.
\newblock Eigen-distortions of hierarchical representations.
\newblock \emph{Advances in Neural Information Processing Systems}, 30, 2017.

\bibitem[Burt \& Adelson(1987)Burt and Adelson]{burt1987laplacian}
Peter~J Burt and Edward~H Adelson.
\newblock The laplacian pyramid as a compact image code.
\newblock In \emph{Readings in computer vision}, pp.\  671--679. Elsevier,
  1987.

\bibitem[Cox et~al.(2005)Cox, Meier, Oertelt, and DiCarlo]{cox2005breaking}
David~D Cox, Philip Meier, Nadja Oertelt, and James~J DiCarlo.
\newblock 'breaking'position-invariant object recognition.
\newblock \emph{Nature neuroscience}, 8\penalty0 (9):\penalty0 1145--1147,
  2005.

\bibitem[Dapello et~al.(2020)Dapello, Marques, Schrimpf, Geiger, Cox, and
  DiCarlo]{dapello2020simulating}
Joel Dapello, Tiago Marques, Martin Schrimpf, Franziska Geiger, David~D Cox,
  and James~J DiCarlo.
\newblock Simulating a primary visual cortex at the front of cnns improves
  robustness to image perturbations.
\newblock \emph{BioRxiv}, 2020.

\bibitem[Deza \& Konkle(2020)Deza and Konkle]{deza2020emergent}
Arturo Deza and Talia Konkle.
\newblock Emergent properties of foveated perceptual systems.
\newblock \emph{arXiv preprint arXiv:2006.07991}, 2020.

\bibitem[Deza et~al.(2019{\natexlab{a}})Deza, Chen, Long, and
  Konkle]{deza2019accelerated}
Arturo Deza, Yi-Chia Chen, Bria Long, and Talia Konkle.
\newblock Accelerated texforms: Alternative methods for generating
  unrecognizable object images with preserved mid-level features.
\newblock In \emph{Conference on Cognitive Computational Neuroscience},
  2019{\natexlab{a}}.

\bibitem[Deza et~al.(2019{\natexlab{b}})Deza, Jonnalagadda, and
  Eckstein]{deza2019towards}
Arturo Deza, Aditya Jonnalagadda, and Miguel~P Eckstein.
\newblock Towards metamerism via foveated style transfer.
\newblock In \emph{International Conference on Learning Representations},
  2019{\natexlab{b}}.

\bibitem[DiCarlo \& Cox(2007)DiCarlo and Cox]{dicarlo2007untangling}
James~J DiCarlo and David~D Cox.
\newblock Untangling invariant object recognition.
\newblock \emph{Trends in cognitive sciences}, 11\penalty0 (8):\penalty0
  333--341, 2007.

\bibitem[Ding et~al.(2020)Ding, Ma, Wang, and Simoncelli]{ding2020image}
K~Ding, K~Ma, S~Wang, and EP~Simoncelli.
\newblock Image quality assessment: Unifying structure and texture similarity.
\newblock \emph{IEEE Transactions on Pattern Analysis and Machine
  Intelligence}, 2020.

\bibitem[Ding et~al.(2021)Ding, Ma, Wang, and Simoncelli]{ding2021comparison}
Keyan Ding, Kede Ma, Shiqi Wang, and Eero~P Simoncelli.
\newblock Comparison of full-reference image quality models for optimization of
  image processing systems.
\newblock \emph{International Journal of Computer Vision}, 129\penalty0
  (4):\penalty0 1258--1281, 2021.

\bibitem[Elsayed et~al.(2018)Elsayed, Shankar, Cheung, Papernot, Kurakin,
  Goodfellow, and Sohl-Dickstein]{elsayed2018adversarial}
Gamaleldin~F Elsayed, Shreya Shankar, Brian Cheung, Nicolas Papernot, Alexey
  Kurakin, Ian~J Goodfellow, and Jascha Sohl-Dickstein.
\newblock Adversarial examples that fool both computer vision and time-limited
  humans.
\newblock In \emph{NeurIPS}, 2018.

\bibitem[Engstrom et~al.(2019)Engstrom, Ilyas, Santurkar, Tsipras, Tran, and
  Madry]{engstrom2019adversarial}
Logan Engstrom, Andrew Ilyas, Shibani Santurkar, Dimitris Tsipras, Brandon
  Tran, and Aleksander Madry.
\newblock Adversarial robustness as a prior for learned representations.
\newblock \emph{arXiv preprint arXiv:1906.00945}, 2019.

\bibitem[Feather et~al.(2019)Feather, Durango, Gonzalez, and
  McDermott]{feather2019metamers}
Jenelle Feather, Alex Durango, Ray Gonzalez, and Josh McDermott.
\newblock Metamers of neural networks reveal divergence from human perceptual
  systems.
\newblock In \emph{Advances in Neural Information Processing Systems}, pp.\
  10078--10089, 2019.

\bibitem[Feather et~al.(2021)Feather, Durango, Leclerc, Madry, and
  McDermott]{feather2021adversarial}
Jenelle Feather, Alex Durango, Guillame Leclerc, Aleksander Madry, and Josh
  McDermott.
\newblock Adversarial training aligns invariances between artificial neural
  networks and biological sensory systems.
\newblock \emph{Cosyne Meeting Abstract}, 2021.

\bibitem[Firestone(2020)]{firestone2020performance}
Chaz Firestone.
\newblock Performance vs. competence in human--machine comparisons.
\newblock \emph{Proceedings of the National Academy of Sciences}, 117\penalty0
  (43):\penalty0 26562--26571, 2020.

\bibitem[Freeman \& Simoncelli(2011)Freeman and
  Simoncelli]{freeman2011metamers}
Jeremy Freeman and Eero~P Simoncelli.
\newblock Metamers of the ventral stream.
\newblock \emph{Nature neuroscience}, 14\penalty0 (9):\penalty0 1195--1201,
  2011.

\bibitem[Funke et~al.(2021)Funke, Borowski, Stosio, Brendel, Wallis, and
  Bethge]{funke2021five}
Christina~M Funke, Judy Borowski, Karolina Stosio, Wieland Brendel, Thomas~SA
  Wallis, and Matthias Bethge.
\newblock Five points to check when comparing visual perception in humans and
  machines.
\newblock \emph{Journal of Vision}, 21\penalty0 (3):\penalty0 16--16, 2021.

\bibitem[Gatys et~al.(2015)Gatys, Ecker, and Bethge]{gatys2015texture}
Leon Gatys, Alexander~S Ecker, and Matthias Bethge.
\newblock Texture synthesis using convolutional neural networks.
\newblock \emph{Advances in neural information processing systems},
  28:\penalty0 262--270, 2015.

\bibitem[Geirhos et~al.(2021)Geirhos, Narayanappa, Mitzkus, Thieringer, Bethge,
  Wichmann, and Brendel]{geirhos2021partial}
Robert Geirhos, Kantharaju Narayanappa, Benjamin Mitzkus, Tizian Thieringer,
  Matthias Bethge, Felix~A Wichmann, and Wieland Brendel.
\newblock Partial success in closing the gap between human and machine vision.
\newblock \emph{Neural Information Processing Systems}, 2021.

\bibitem[Geisler \& Perry(1998)Geisler and Perry]{geisler1998real}
Wilson~S Geisler and Jeffrey~S Perry.
\newblock Real-time foveated multiresolution system for low-bandwidth video
  communication.
\newblock In \emph{Human vision and electronic imaging III}, volume 3299, pp.\
  294--305. International Society for Optics and Photonics, 1998.

\bibitem[Gilmer et~al.(2018)Gilmer, Metz, Faghri, Schoenholz, Raghu,
  Wattenberg, and Goodfellow]{gilmer2018adversarial}
Justin Gilmer, Luke Metz, Fartash Faghri, Sam Schoenholz, Maithra Raghu, Martin
  Wattenberg, and Ian Goodfellow.
\newblock Adversarial spheres, 2018.
\newblock URL \url{https://openreview.net/forum?id=SyUkxxZ0b}.

\bibitem[Golan et~al.(2020)Golan, Raju, and
  Kriegeskorte]{golan2019controversial}
Tal Golan, Prashant~C. Raju, and Nikolaus Kriegeskorte.
\newblock Controversial stimuli: Pitting neural networks against each other as
  models of human cognition.
\newblock \emph{Proceedings of the National Academy of Sciences}, 117\penalty0
  (47):\penalty0 29330--29337, 2020.
\newblock ISSN 0027-8424.
\newblock \doi{10.1073/pnas.1912334117}.
\newblock URL \url{https://www.pnas.org/content/117/47/29330}.

\bibitem[Goodfellow et~al.(2015)Goodfellow, Shlens, and
  Szegedy]{goodfellow2014explaining}
Ian~J Goodfellow, Jonathon Shlens, and Christian Szegedy.
\newblock Explaining and harnessing adversarial examples.
\newblock \emph{International Conference on Learning Representations}, 2015.

\bibitem[Gowal et~al.(2021)Gowal, Rebuffi, Wiles, Stimberg, Calian, and
  Mann]{gowal2021improving}
Sven Gowal, Sylvestre-Alvise Rebuffi, Olivia Wiles, Florian Stimberg,
  Dan~Andrei Calian, and Timothy~A Mann.
\newblock Improving robustness using generated data.
\newblock \emph{Advances in Neural Information Processing Systems}, 34, 2021.

\bibitem[Han et~al.(2020)Han, Roig, Geiger, and Poggio]{han2020scale}
Yena Han, Gemma Roig, Gad Geiger, and Tomaso Poggio.
\newblock Scale and translation-invariance for novel objects in human vision.
\newblock \emph{Scientific reports}, 10\penalty0 (1):\penalty0 1--13, 2020.

\bibitem[Herrera-Esposito et~al.(2021)Herrera-Esposito, Coen-Cagli, and
  Gomez-Sena]{herrera2021flexible}
Daniel Herrera-Esposito, Ruben Coen-Cagli, and Leonel Gomez-Sena.
\newblock Flexible contextual modulation of naturalistic texture perception in
  peripheral vision.
\newblock \emph{Journal of vision}, 21\penalty0 (1):\penalty0 1--1, 2021.

\bibitem[Hinton(2021)]{hinton2021represent}
Geoffrey Hinton.
\newblock How to represent part-whole hierarchies in a neural network.
\newblock \emph{arXiv preprint arXiv:2102.12627}, 2021.

\bibitem[Ilyas et~al.(2019)Ilyas, Santurkar, Tsipras, Engstrom, Tran, and
  Madry]{ilyas2019adversarial}
Andrew Ilyas, Shibani Santurkar, Dimitris Tsipras, Logan Engstrom, Brandon
  Tran, and Aleksander Madry.
\newblock Adversarial examples are not bugs, they are features.
\newblock In \emph{Advances in Neural Information Processing Systems}, pp.\
  125--136, 2019.

\bibitem[Jonnalagadda et~al.(2021)Jonnalagadda, Wang, and
  Eckstein]{jonnalagadda2021foveater}
Aditya Jonnalagadda, William Wang, and Miguel~P Eckstein.
\newblock Foveater: Foveated transformer for image classification.
\newblock \emph{arXiv preprint arXiv:2105.14173}, 2021.

\bibitem[LeCun et~al.(2015)LeCun, Bengio, and Hinton]{lecun2015deep}
Yann LeCun, Yoshua Bengio, and Geoffrey Hinton.
\newblock Deep learning.
\newblock \emph{nature}, 521\penalty0 (7553):\penalty0 436--444, 2015.

\bibitem[Liu et~al.(2016)Liu, Gousseau, and Xia]{liu2016texture}
Gang Liu, Yann Gousseau, and Gui-Song Xia.
\newblock Texture synthesis through convolutional neural networks and spectrum
  constraints.
\newblock In \emph{2016 23rd International Conference on Pattern Recognition
  (ICPR)}, pp.\  3234--3239. IEEE, 2016.

\bibitem[Logothetis et~al.(1995)Logothetis, Pauls, and
  Poggio]{logothetis1995shape}
Nikos~K Logothetis, Jon Pauls, and Tomaso Poggio.
\newblock Shape representation in the inferior temporal cortex of monkeys.
\newblock \emph{Current biology}, 5\penalty0 (5):\penalty0 552--563, 1995.

\bibitem[Long et~al.(2018)Long, Yu, and Konkle]{long2018mid}
Bria Long, Chen-Ping Yu, and Talia Konkle.
\newblock Mid-level visual features underlie the high-level categorical
  organization of the ventral stream.
\newblock \emph{Proceedings of the National Academy of Sciences}, 115\penalty0
  (38):\penalty0 E9015--E9024, 2018.

\bibitem[Madry et~al.(2017)Madry, Makelov, Schmidt, Tsipras, and
  Vladu]{madry2017towards}
Aleksander Madry, Aleksandar Makelov, Ludwig Schmidt, Dimitris Tsipras, and
  Adrian Vladu.
\newblock Towards deep learning models resistant to adversarial attacks.
\newblock \emph{International Conference on Learning Representations}, 2017.

\bibitem[Mahendran \& Vedaldi(2015)Mahendran and
  Vedaldi]{mahendran2015understanding}
Aravindh Mahendran and Andrea Vedaldi.
\newblock Understanding deep image representations by inverting them.
\newblock In \emph{Proceedings of the IEEE conference on computer vision and
  pattern recognition}, pp.\  5188--5196, 2015.

\bibitem[Martinez-Garcia et~al.(2019)Martinez-Garcia, Bertalm{\'\i}o, and
  Malo]{martinez2019praise}
Marina Martinez-Garcia, Marcelo Bertalm{\'\i}o, and Jes{\'u}s Malo.
\newblock In praise of artifice reloaded: Caution with natural image databases
  in modeling vision.
\newblock \emph{Frontiers in neuroscience}, 13:\penalty0 8, 2019.

\bibitem[Nandy \& Tjan(2012)Nandy and Tjan]{nandy2012saccade}
Anirvan~S Nandy and Bosco~S Tjan.
\newblock Saccade-confounded image statistics explain visual crowding.
\newblock \emph{Nature neuroscience}, 15\penalty0 (3):\penalty0 463--469, 2012.

\bibitem[Poggio et~al.(2014)Poggio, Mutch, and Isik]{poggio2014computational}
Tomaso Poggio, Jim Mutch, and Leyla Isik.
\newblock Computational role of eccentricity dependent cortical magnification.
\newblock \emph{arXiv preprint arXiv:1406.1770}, 2014.

\bibitem[Portilla \& Simoncelli(2000)Portilla and
  Simoncelli]{portilla2000parametric}
Javier Portilla and Eero~P Simoncelli.
\newblock A parametric texture model based on joint statistics of complex
  wavelet coefficients.
\newblock \emph{International journal of computer vision}, 40\penalty0
  (1):\penalty0 49--70, 2000.

\bibitem[Rebuffi et~al.(2021)Rebuffi, Gowal, Calian, Stimberg, Wiles, and
  Mann]{rebuffi2021fixing}
Sylvestre-Alvise Rebuffi, Sven Gowal, Dan~A Calian, Florian Stimberg, Olivia
  Wiles, and Timothy Mann.
\newblock Fixing data augmentation to improve adversarial robustness.
\newblock \emph{Neural Information Processing Systems}, 2021.

\bibitem[Reddy et~al.(2020)Reddy, Banburski, Pant, and
  Poggio]{reddy2020biologically}
Manish~V Reddy, Andrzej Banburski, Nishka Pant, and Tomaso Poggio.
\newblock Biologically inspired mechanisms for adversarial robustness.
\newblock \emph{Neural Information Processing Systems}, 2020.

\bibitem[Riesenhuber \& Poggio(1999)Riesenhuber and
  Poggio]{riesenhuber1999hierarchical}
Maximilian Riesenhuber and Tomaso Poggio.
\newblock Hierarchical models of object recognition in cortex.
\newblock \emph{Nature neuroscience}, 2\penalty0 (11):\penalty0 1019--1025,
  1999.

\bibitem[Rosenholtz(2016)]{rosenholtz2016capabilities}
Ruth Rosenholtz.
\newblock Capabilities and limitations of peripheral vision.
\newblock \emph{Annual Review of Vision Science}, 2:\penalty0 437--457, 2016.

\bibitem[Rosenholtz et~al.(2012)Rosenholtz, Huang, Raj, Balas, and
  Ilie]{rosenholtz2012summary}
Ruth Rosenholtz, Jie Huang, Alvin Raj, Benjamin~J Balas, and Livia Ilie.
\newblock A summary statistic representation in peripheral vision explains
  visual search.
\newblock \emph{Journal of vision}, 12\penalty0 (4):\penalty0 14--14, 2012.

\bibitem[Russakovsky et~al.(2015)Russakovsky, Deng, Su, Krause, Satheesh, Ma,
  Huang, Karpathy, Khosla, Bernstein, et~al.]{russakovsky2015imagenet}
Olga Russakovsky, Jia Deng, Hao Su, Jonathan Krause, Sanjeev Satheesh, Sean Ma,
  Zhiheng Huang, Andrej Karpathy, Aditya Khosla, Michael Bernstein, et~al.
\newblock Imagenet large scale visual recognition challenge.
\newblock \emph{International journal of computer vision}, 115\penalty0
  (3):\penalty0 211--252, 2015.

\bibitem[Santurkar et~al.(2019)Santurkar, Tsipras, Tran, Ilyas, Engstrom, and
  Madry]{santurkar2019image}
Shibani Santurkar, Dimitris Tsipras, Brandon Tran, Andrew Ilyas, Logan
  Engstrom, and Aleksander Madry.
\newblock Image synthesis with a single (robust) classifier.
\newblock \emph{Neural Information Processing Systems}, 2019.

\bibitem[Schmidhuber(2015)]{schmidhuber2015deep}
J{\"u}rgen Schmidhuber.
\newblock Deep learning in neural networks: An overview.
\newblock \emph{Neural networks}, 61:\penalty0 85--117, 2015.

\bibitem[Stewart et~al.(2020)Stewart, Valsecchi, and
  Sch{\"u}tz]{stewart2020review}
Emma~EM Stewart, Matteo Valsecchi, and Alexander~C Sch{\"u}tz.
\newblock A review of interactions between peripheral and foveal vision.
\newblock \emph{Journal of vision}, 20\penalty0 (12):\penalty0 2--2, 2020.

\bibitem[Szegedy et~al.(2014)Szegedy, Zaremba, Sutskever, Bruna, Erhan,
  Goodfellow, and Fergus]{szegedy2013intriguing}
Christian Szegedy, Wojciech Zaremba, Ilya Sutskever, Joan Bruna, Dumitru Erhan,
  Ian Goodfellow, and Rob Fergus.
\newblock Intriguing properties of neural networks.
\newblock \emph{International Conference on Learning Representations}, 2014.

\bibitem[Tsipras et~al.(2019)Tsipras, Santurkar, Engstrom, Turner, and
  Madry]{tsipras2018robustness}
Dimitris Tsipras, Shibani Santurkar, Logan Engstrom, Alexander Turner, and
  Aleksander Madry.
\newblock Robustness may be at odds with accuracy.
\newblock \emph{International Conference on Learning Representations}, 2019.

\bibitem[Van~der Maaten \& Hinton(2008)Van~der Maaten and
  Hinton]{van2008visualizing}
Laurens Van~der Maaten and Geoffrey Hinton.
\newblock Visualizing data using t-sne.
\newblock \emph{Journal of machine learning research}, 9\penalty0 (11), 2008.

\bibitem[Von~Helmholtz(1867)]{von1867handbuch}
Hermann Von~Helmholtz.
\newblock \emph{Handbuch der physiologischen Optik: mit 213 in den Text
  eingedruckten Holzschnitten und 11 Tafeln}, volume~9.
\newblock Voss, 1867.

\bibitem[Wallis et~al.(2016)Wallis, Bethge, and Wichmann]{wallis2016testing}
Thomas~SA Wallis, Matthias Bethge, and Felix~A Wichmann.
\newblock Testing models of peripheral encoding using metamerism in an oddity
  paradigm.
\newblock \emph{Journal of vision}, 16\penalty0 (2):\penalty0 4--4, 2016.

\bibitem[Wallis et~al.(2017)Wallis, Funke, Ecker, Gatys, Wichmann, and
  Bethge]{wallis2017parametric}
Thomas~SA Wallis, Christina~M Funke, Alexander~S Ecker, Leon~A Gatys, Felix~A
  Wichmann, and Matthias Bethge.
\newblock A parametric texture model based on deep convolutional features
  closely matches texture appearance for humans.
\newblock \emph{Journal of vision}, 17\penalty0 (12):\penalty0 5--5, 2017.

\bibitem[Wallis et~al.(2019)Wallis, Funke, Ecker, Gatys, Wichmann, and
  Bethge]{wallis2019image}
Thomas~SA Wallis, Christina~M Funke, Alexander~S Ecker, Leon~A Gatys, Felix~A
  Wichmann, and Matthias Bethge.
\newblock Image content is more important than bouma’s law for scene
  metamers.
\newblock \emph{ELife}, 8:\penalty0 e42512, 2019.

\bibitem[Williams et~al.(2008)Williams, Baker, De~Beeck, Shim, Dang,
  Triantafyllou, and Kanwisher]{williams2008feedback}
Mark~A Williams, Chris~I Baker, Hans P~Op De~Beeck, Won~Mok Shim, Sabin Dang,
  Christina Triantafyllou, and Nancy Kanwisher.
\newblock Feedback of visual object information to foveal retinotopic cortex.
\newblock \emph{Nature neuroscience}, 11\penalty0 (12):\penalty0 1439--1445,
  2008.

\bibitem[Ziemba \& Simoncelli(2021)Ziemba and Simoncelli]{ziemba2021opposing}
Corey~M Ziemba and Eero~P Simoncelli.
\newblock Opposing effects of selectivity and invariance in peripheral vision.
\newblock \emph{Nature Communications}, 12\penalty0 (1):\penalty0 1--11, 2021.

\bibitem[Ziemba et~al.(2016)Ziemba, Freeman, Movshon, and
  Simoncelli]{ziemba2016selectivity}
Corey~M Ziemba, Jeremy Freeman, J~Anthony Movshon, and Eero~P Simoncelli.
\newblock Selectivity and tolerance for visual texture in macaque v2.
\newblock \emph{Proceedings of the National Academy of Sciences}, 113\penalty0
  (22):\penalty0 E3140--E3149, 2016.

\end{thebibliography}
\bibliographystyle{iclr2022_conference}

\clearpage
\newpage

\appendix







\section{Image Synthesis Details}

\begin{table}[h!]
\footnotesize
\setlength{\tabcolsep}{5pt}
\renewcommand{\arraystretch}{1.25}
\centering
 \begin{tabular}{c| c c c c c c c c c c c c} 
 \hline
 \multicolumn{10}{c}{\textbf{Classes}}\\
 \hline
 \textbf{RIN} & Dog & Cat & Frog & Turtle & Bird & Primate & Fish & Crab & Insect \\
 \hline
 \textbf{IN} & 151-268 & 281-285 & 30-32 & 33-37 & 68-100 & 365-382 & 389-397 & 118-121 & 300-319 \\
 \hline
 \end{tabular}
 \caption{Classes of RestrictedImageNet (\textbf{RIN}) and the corresponding ImageNet (\textbf{IN}) class ranges.}
 \label{data_classes}
\end{table}

\subsection{Standard and Robust Stimuli}

\begin{figure}
    \centering
    \includegraphics[scale=0.7]{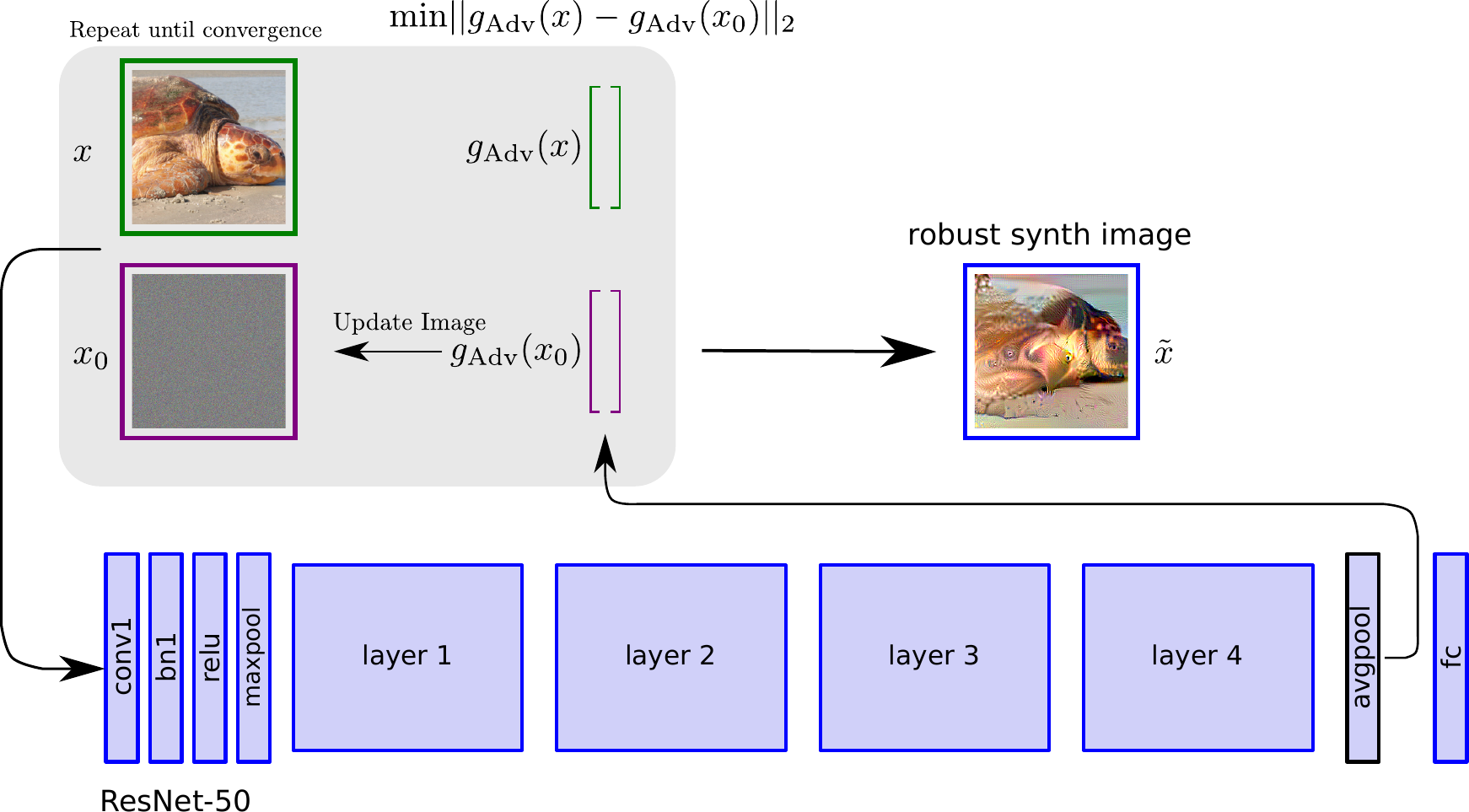}
    \caption{The Robust Image Synthesis pipeline: A noise image $x_0$ is passed through an adversarially trained ResNet-50 and the penultimate layer features $g_{\text{Adv}}(x_0)$ are matched wrt the original images' penultimate feature activation $g_{\text{Adv}}(x)$ via an L2 loss, and is repeated until convergence~\citep{santurkar2019image,engstrom2019adversarial}. Critically we use $g_{\text{Adv}}(\circ)$ as a summary statistic of peripheral processing in our experiments. } 
    \label{fig:synthesis_prodecure}
\end{figure}

We used the publicly available code from~\cite{santurkar2019image,engstrom2019adversarial,ilyas2019adversarial} found here to synthesize both standard and robust stimuli which where derived from a regularly and adversarially trained model respectively:
\url{https://github.com/MadryLab/robust_representations}

A schematic that illustrates the robust stimuli rendering pipeline can be seen in Figure~\ref{fig:synthesis_prodecure}. Standard stimuli is generated with the same procedure, and number of iterations, but the network $g_\text{Adv}(\circ)$ is replaced with $g_\text{Standard}(\circ)$ instead.

\begin{figure}
    \centering
    \includegraphics[width=\columnwidth]{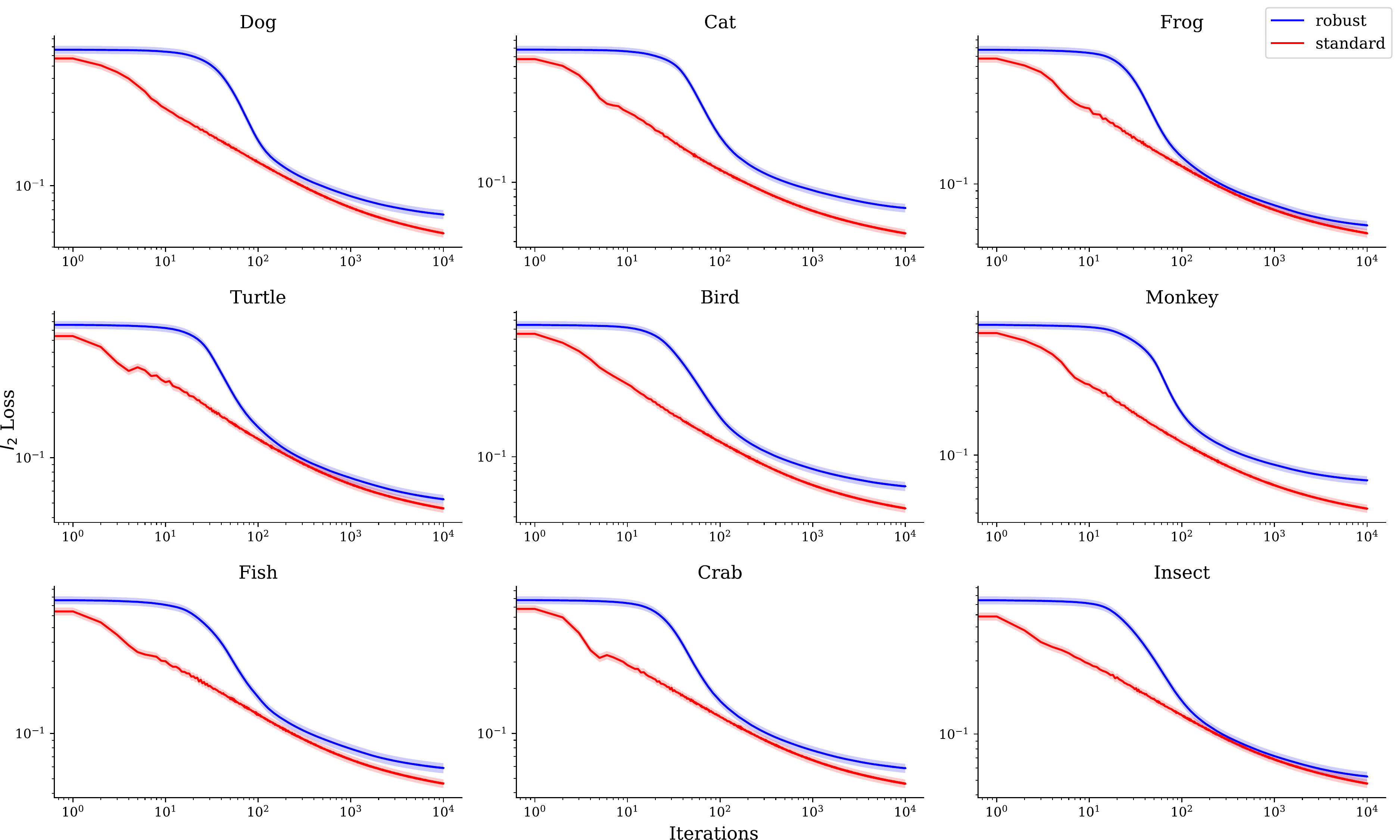}
    \caption{Per-Class Synthesis Loss visualizations for the Robust and Standard Stimuli across all samples. Errorbar represents 1 standard error.} 
    \label{fig:synthesis_loss}
\end{figure}

A visualization of the convergence of the loss when performing the synthesis procedure can be seen in Figure~\ref{fig:synthesis_loss}.

\subsection{Texform Stimuli}
\label{sec:Sup_Texforms}

Texform stimuli were synthesized using the publicly available code of~\cite{deza2019accelerated}: \url{https://github.com/ArturoDeza/Fast-Texforms}

The following images (class:[image id's]) were removed as they did not converge:

\begin{itemize}
\item texform0: {0:[49],1:[9],2:[],3:[44],4:[],5:[],6:[10],7:[40],8:[]}.
\item texform1: {0:[49],1:[9,44],2:[],3:[44],4:[],5:[],6:[10],7:[40],8:[]}
\end{itemize}

\begin{figure}
    \centering
    \includegraphics[width=\columnwidth]{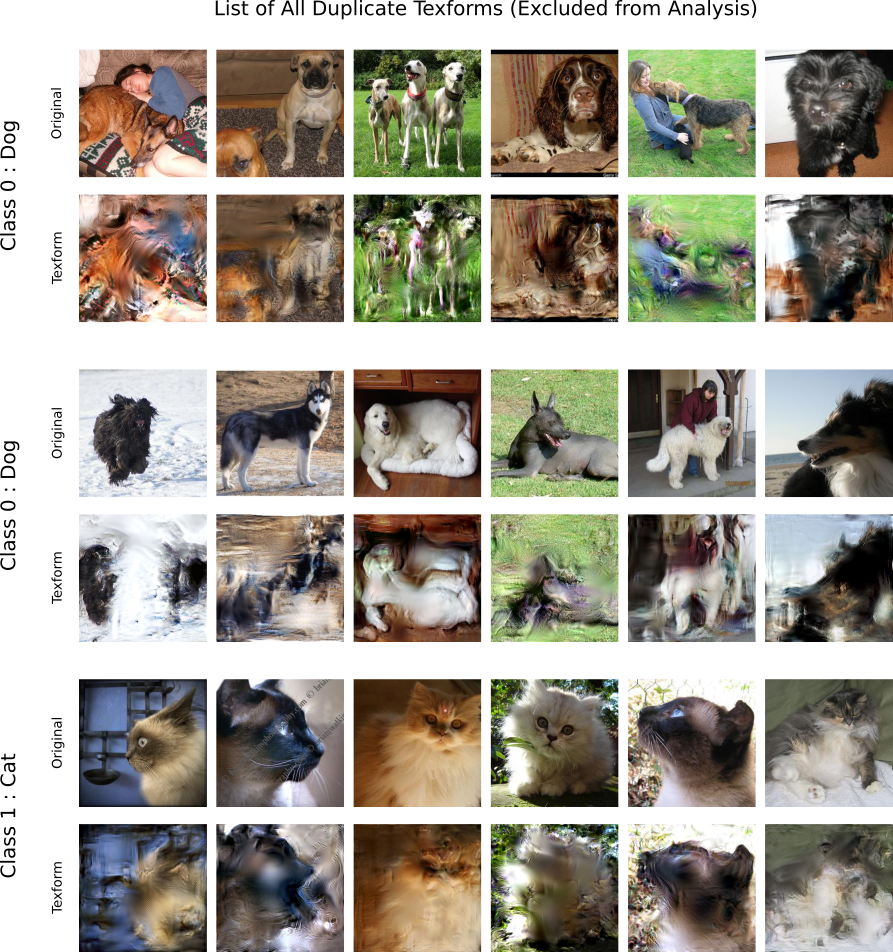}
    \caption{Duplicates are images that even though they were initialized with two different random noise images, they converged to the exact same image (Mean Square Error between synthesized samples is equal to zero). These stimuli were \textit{excluded} from our analysis and represent only $2\%$ $(18/894)$ of the used texform stimuli.}
    \label{fig:Duplicate_Texforms}
\end{figure}

In addition the following image id's were removed from our psychophysical analysis from the texform stimuli as they converged to the \textit{exact} same image even when starting from different noise seeds. This was found while doing a post-hoc IQA analysis as the one shown in Figure~\ref{fig:Duplicate_Texforms}. These stimuli only occurred for classes 0 (dog) and 1 (cat):

\begin{itemize}
    \item texform: {0:[22,25,26,27,29,93,94,95,96,97,98,99],1:[20,21,22,23,73,74]}
\end{itemize}

We found that Standard and Robust stimuli did not have this identical convergence problem over the 900 rendered pairs (1800 stimuli in total for Standard and 1800 in total for Robust).

\textbf{Note 1a:} A common mis-conception is that~\cite{freeman2011metamers}-derived stimuli (such as texforms) \textit{do not} contain structural priors and only performs localized texture synthesis over smoothly overlapping log-polar receptive fields. This has been investigated with great detail in~\cite{wallis2016testing,wallis2017parametric,liu2016texture} that showed that without spectral constraints it is impossible to generate metameric images from non-stationary textures for the human observer when showing such stimuli in the visual periphery. For texforms the metameric constraint is purposely broken because we'd like to test how a specific biologically-plausible family of transformations (embodied through the synthesis procedure) interacts with eccentricity when the eccentricity-dependent and scaling factors texform parameters are fixed. See $(z_*,s_*)$ from Eq.~\ref{eq:Texform_Optimization}. 

\textbf{Note 1b:} The \cite{freeman2011metamers} synthesis model is not equivalent to the~\cite{portilla2000parametric} synthesis model. The \cite{freeman2011metamers} is a super-ordinate synthesis model class that locally uses the \cite{portilla2000parametric} synthesis model over smoothly overlapping receptive fields in addition to adding a global structural prior. Texforms are rendered with the \cite{freeman2011metamers} model, by placing he simulated point of fixation \textit{outside} the image~\citep{long2018mid,deza2019accelerated}.

\textbf{Note 1c:} Usual texform rendering time is about 1 day per image, though the rendering procedure has been accelerated to the order of minutes as shown in~\cite{deza2019accelerated}. We used their publicly available code in our experiments. Thus, it is worth noting that synthesizing texforms in the order of hundreds of thousands (or millions) for supervised learning experiments -- has not been done before and is computationally expensive (may take months), which is why Figure~\ref{fig:Stimuli_Collection} displays no information on texform-trained CNN's. This direction is current work.

\textbf{Note 2:} A first naive criticism to the selection of making texforms fixed and not varying as a function of eccentricity -- given the model they were based on~\citep{freeman2011metamers} -- is that they will not create metameric stimuli. Our anticipated reply to this is three-fold, and partially aligned with the motivation of~\cite{long2018mid}:

\begin{enumerate}
\item Our goal is \textit{not} to make metameric stimuli out of texforms or robust stimuli, but to examine how perceptual discriminability rates of a \textit{fixed stimuli} change as a function of retinal eccentricity. By checking if these perceptual decays are similar (which we show) we can connect both functions that give rise to these apparently un-related transformations (the stimuli). Recall Eq.~\ref{eq:Psychometric_Equals}.
\item Having a ``metameric texform'' that changes as a function of eccentricity would defeat the purpose of using it as a control in our experiments. Had this been the road taken, we would now have a control curve that will presumably be horizontal and at chance, providing no information about how the transformation that gives rise to the robust stimuli is linked to the texform transformation.
\item The goal of this paper is \textit{not} to make a foveated metamer model that fools human observers similar to that of~\cite{freeman2011metamers,rosenholtz2012summary,deza2019towards,wallis2019image} that would be based on a foveated adversarially trained network. The previous idea however is highly interesting and is being explored in current work, and this work provides a proof of concept that it is tractable.
\end{enumerate}

\subsection{Synthesis vs Synthesis and Original vs Original}

The goal of combining these experimental variations into a block (called \textit{`stimulus roving'}) in our experiments was two-fold: 1) to add difficulty to the tasks thus reducing the likelihood of ceiling effects; 2) to gather two psychometric functions per family of stimuli, which portrays a better description of each stimulus's evoked perceptual signatures. Synthesis vs Synthesis experiments probe the diversity of samples in pixel space that can potentially yield visual metamerism, while the Original vs Synthesis condition yields a stronger condition for visual metamerism. Several works have explored these paradigms~\citep{wallis2016testing,deza2019towards}.

\clearpage
\newpage

\subsection{Sample Stimuli}

\begin{figure}[!h]
    \centering
    \includegraphics[width=\columnwidth]{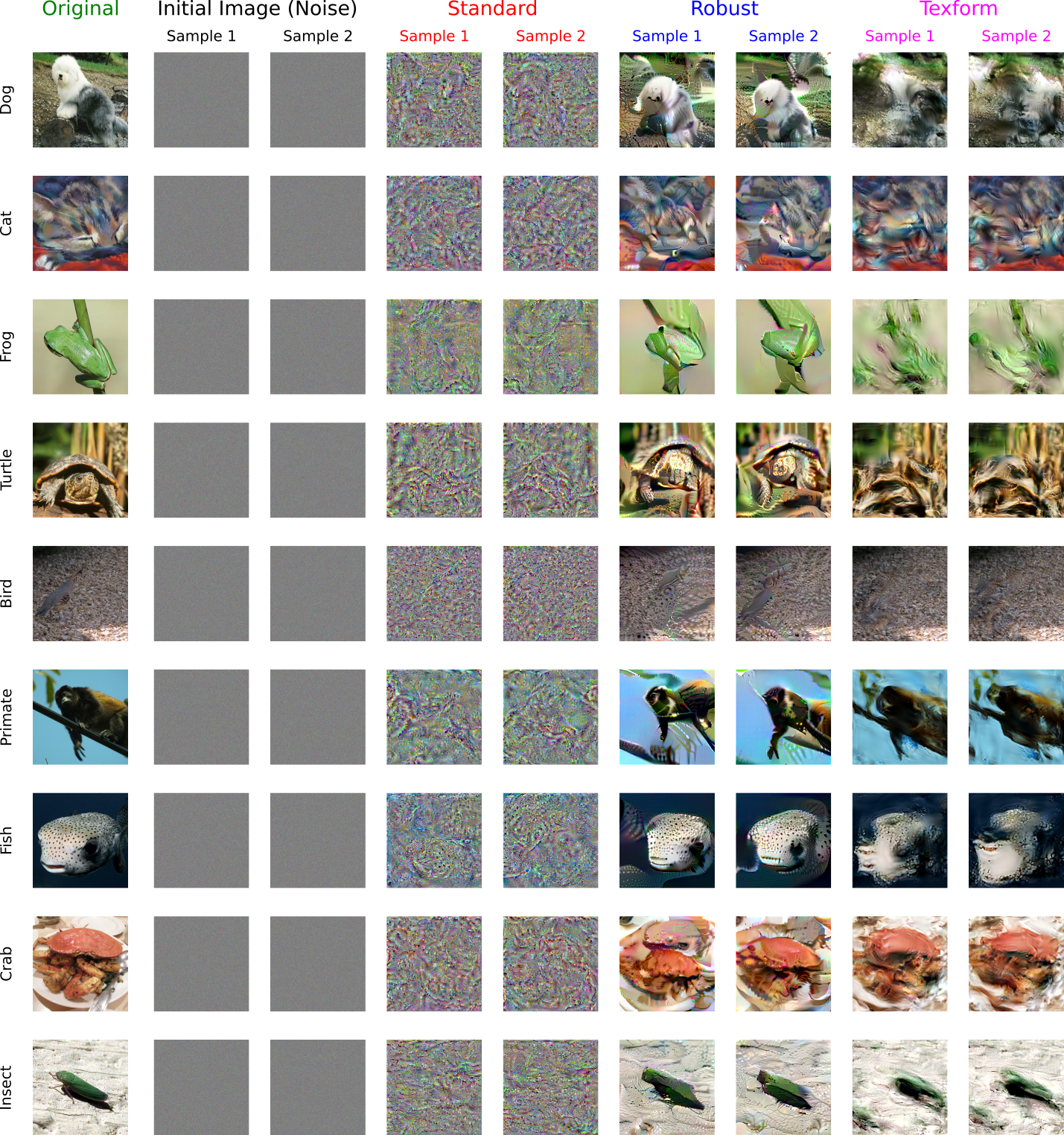}
    \caption{A collection of sample stimuli for each image class used in our experiments.}
    \label{fig:Sample_Stimuli}
\end{figure}

\clearpage
\newpage

\subsection{Synthesis Variations}

In this sub-section we show a collection of different synthesized samples using different reference images, and also using different starting images. If the transformations undergoing the texform model and the adversarially robust model are similar, then the resulting synthesis outputs should look similar if the output of one model is used as the input to another (See inset 2). A similar effect should occur if the starting image for the texform model is robust stimuli and vice-versa (See inset 3). We see this effects qualitatively holds even more so for the Turtle than the Cat image. Overall there are striking low-frequency structural similarities across all images in the last 2 columns. However, further psychophysical experiments are needed to test the rates of discriminability of such images as a function of retinal eccentricity to establish a more precise relationship between them.

\begin{figure}[!h]
    \centering
    \includegraphics[width=0.8\columnwidth]{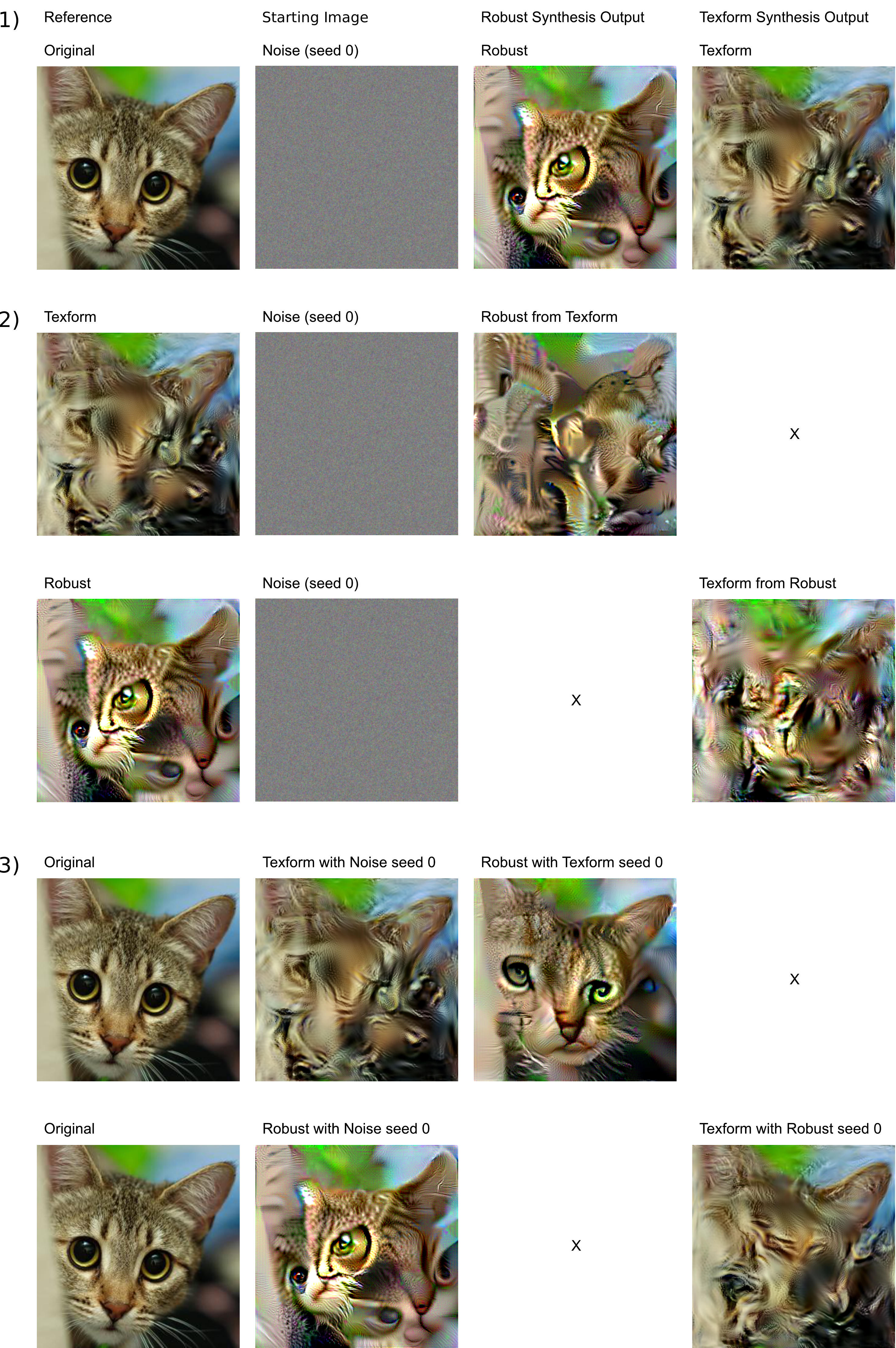}
    \caption{Robust and Texform synthesis variations of a cat with different reference images and starting images.}
    \label{fig:Synthesis_Variation_Cat}
\end{figure}

\clearpage
\newpage

\begin{figure}[!h]
    \centering
    \includegraphics[width=0.8\columnwidth]{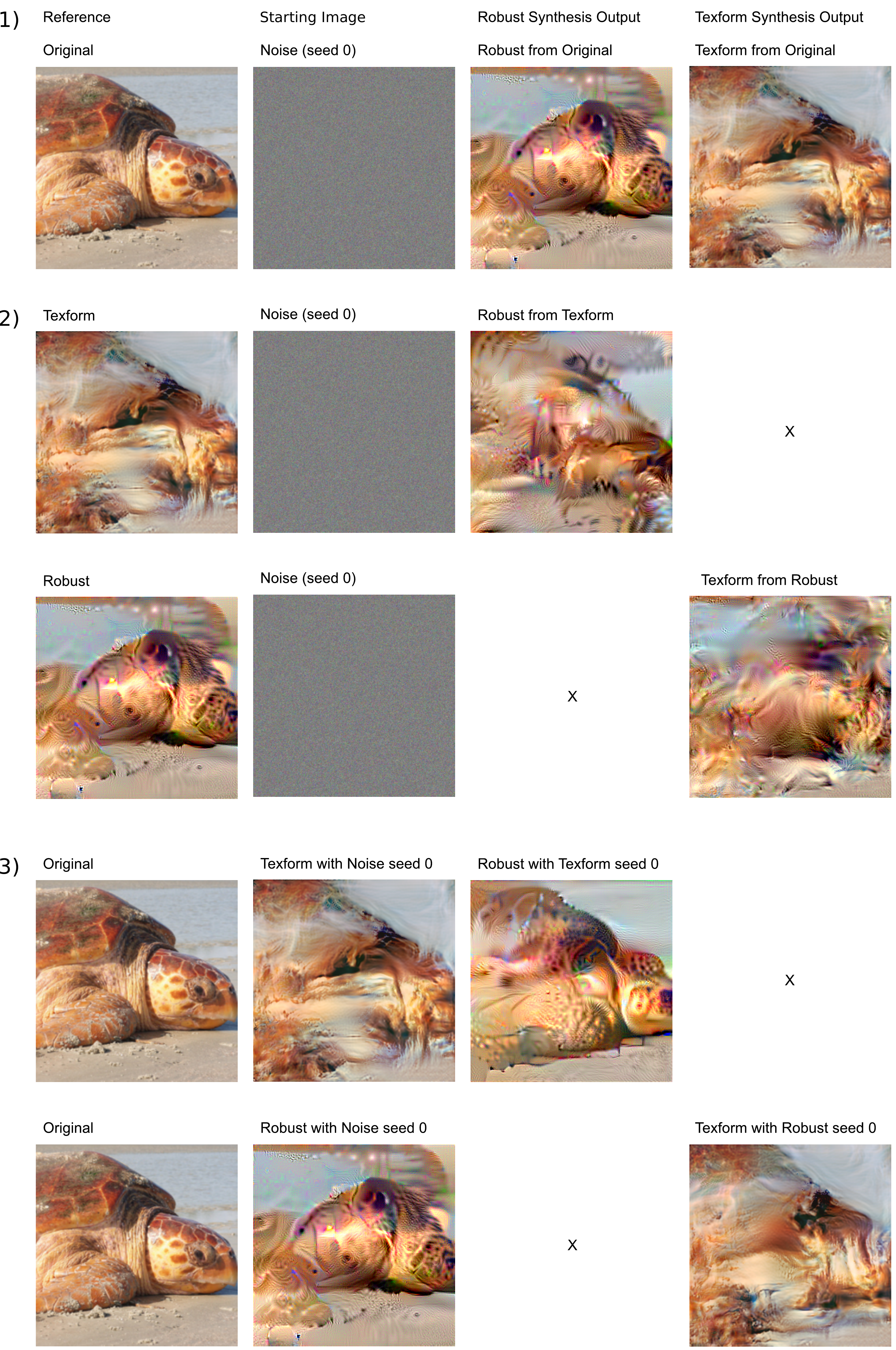}
    \caption{Robust and Texform synthesis variations of a turtle with different reference images and starting images.}
    \label{fig:Synthesis_Variation_Turtle}
\end{figure}

\clearpage
\newpage

\subsection{Ethics Statement, Additional Methods \& Single Observer Results}
\label{sec:Apparatus}

\begin{figure}[!b]
    \centering
    \includegraphics[width=1.0\columnwidth]{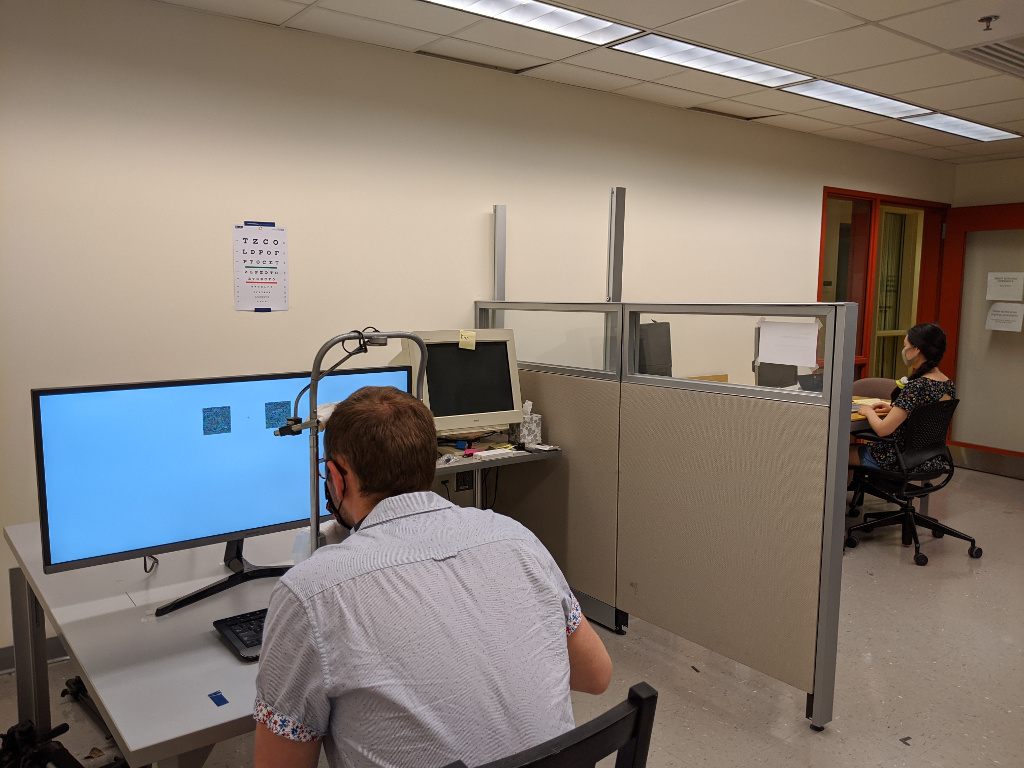}
    \caption{A visualization of how the psychophysical experiments were ran.}
    \label{fig:Psychophysics_Demo}
\end{figure}

All 12 human subjects involved in this research willfully participated in this experiment via explicit consent during each session of this experiment. Our experimental design was reviewed and approved by an Institutional Review Board (IRB).

\textbf{Subjects}: We used a total of 12 human subjects that consisted of undergraduates from the Massachusetts Institute of Technology. Subjects were paid a fee of $\$20$ per hour to complete the experiment in a total of 6 hours over anywhere between 2 to 6 days where observers performed a maximum of 2 hours of psychophysics per day. Our experiments had an approved IRB protocol from the Massachusetts Institute of Technology. Human participants were all tested with a Snellen eye-chart and had at least 20/20 visual acuity and had either no visual correction or contact lenses to correct their vision. All participants were naive to the experiment (\textit{i.e.} no participants were the experimenters), in all cases, subjects were not familiar with the concepts of either visual metamerism or adversarial images. No participants with eye-glasses were used in our experiments.

\textbf{Apparatus}: Experiments were ran on an Ubuntu-Linux Machine version 14.04.5 LTS with MATLAB 2015a's Psychtoolbox version 3.0.14. A chin rest was used so that observers can view stimuli on a screen placed at 50 cm distance from their eyes. We used a 34 inch diagonal 75 Hz LCD monitor, that measured 80cm width and 34 cm height with a visual display resolution of 3440 pixels width by 1440 pixels height. From here the total degrees of visual angle was computed via:

\begin{equation}
    \theta = 2\times\text{atan}(17.0/50.0)\times180.0/\pi
\end{equation}

And the degrees of visual angle subtended by the stimuli is computed by multiplicating the proportion of pixels subtended by the stimuli with respect to the monitor:

\begin{equation}
    \theta_\text{Stimuli} = \theta \times 256/1440 = 6.67
\end{equation}

In the rest of this sub-section we plot the single observer results where the trends observed in Figure~\ref{fig:summary_fig_1} still hold true at the individual per-observer level.  Each participant saw 72 trials of the oddity task for every stimuli condition and eccentricity (i.e. robust synthesized vs synthesized at 5 degrees, robust synthesized vs original at 5 degrees, etc ...). On the 2AFC matching, they saw 80. Errorbars in each plot were computed via a 10,000 sample bootstrapping and represent the 95\% confidence interval.

\newpage

\begin{figure}
    \centering
    \includegraphics[width=1.0\columnwidth]{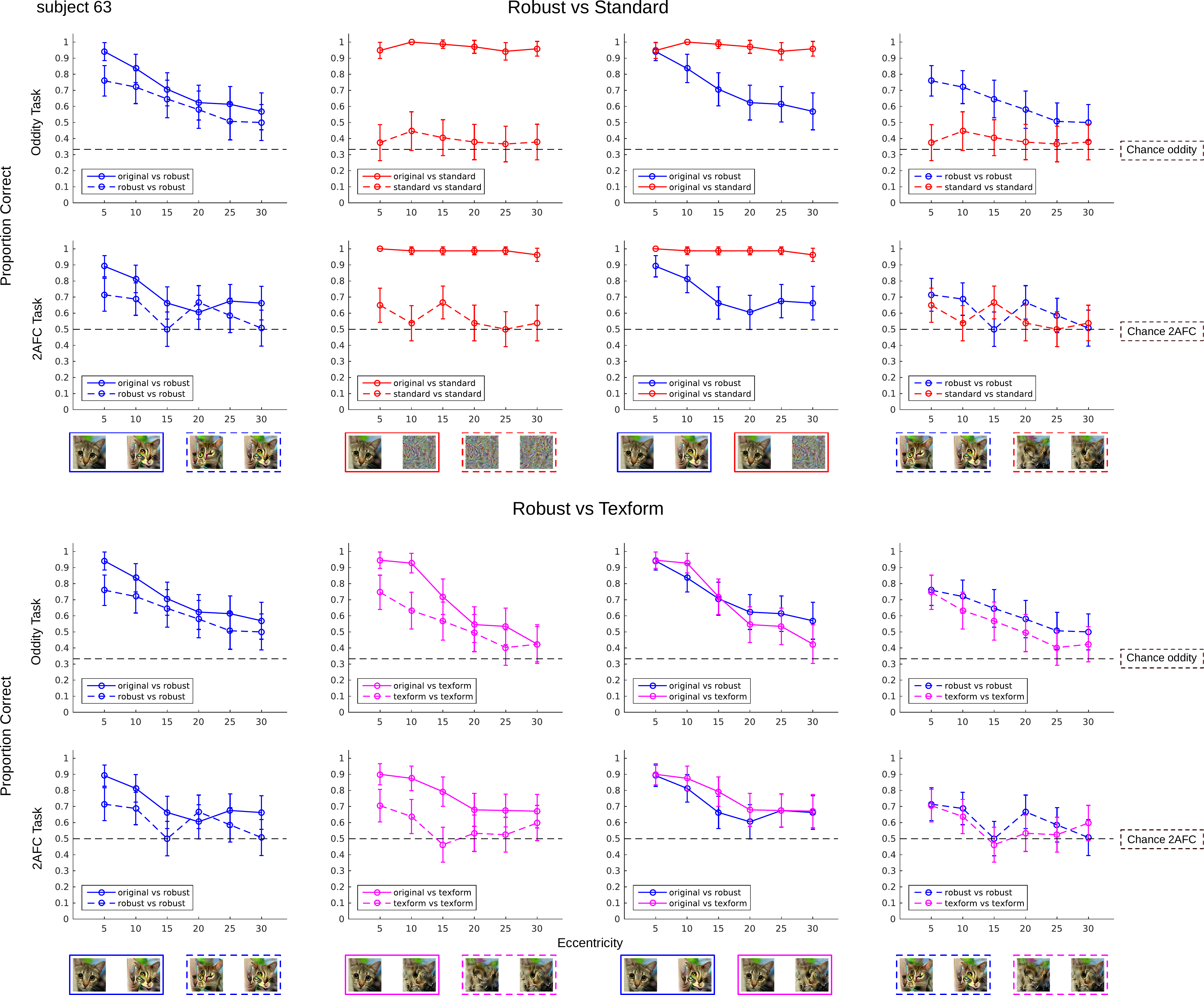}
    \caption{Subject 63}
    \label{fig:subject_JH63}
\end{figure}

\begin{figure}
    \centering
    \includegraphics[width=1.0\columnwidth]{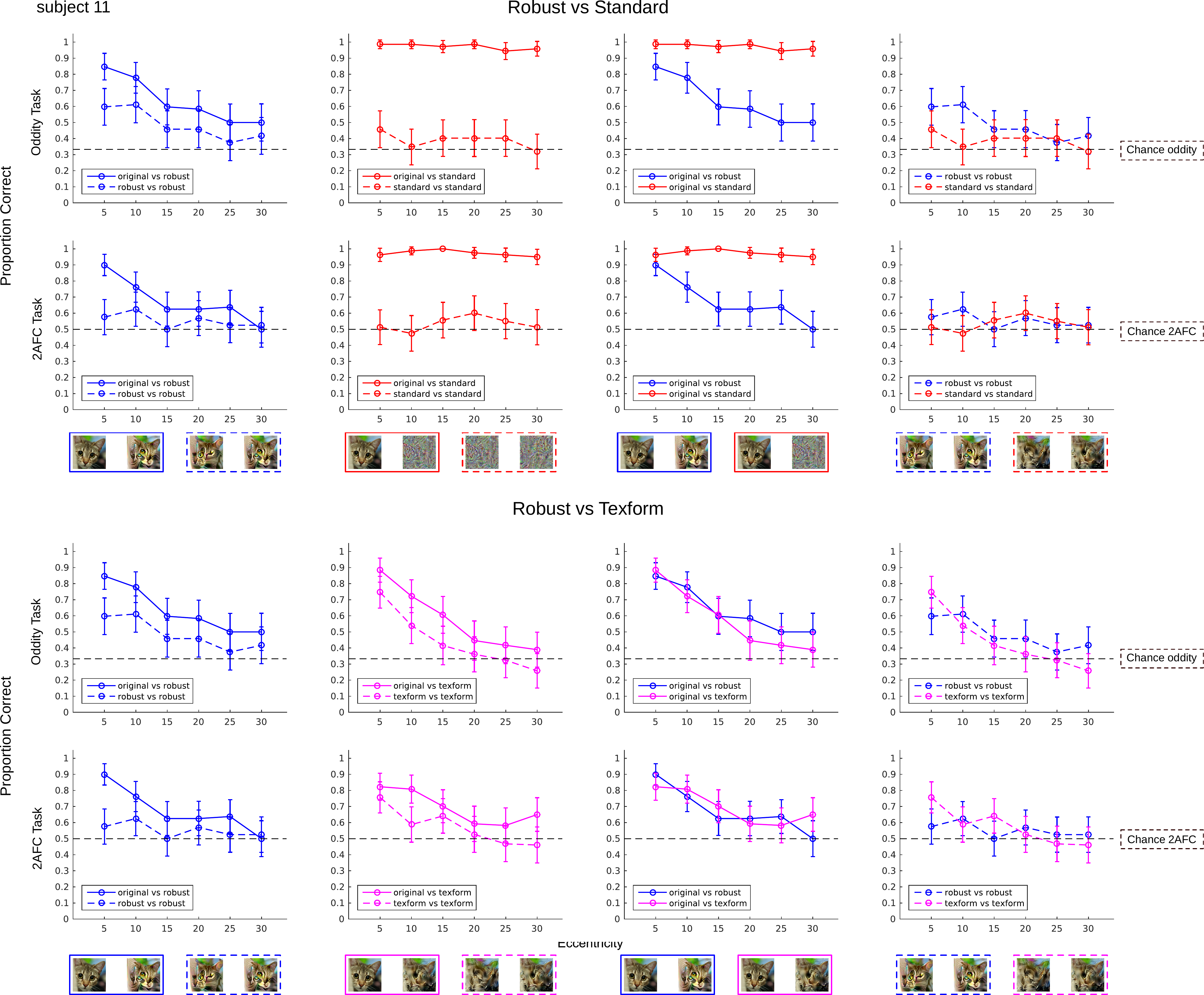}
    \caption{Subject 11}
    \label{fig:subject_11}
\end{figure}

\begin{figure}
    \centering
    \includegraphics[width=1.0\columnwidth]{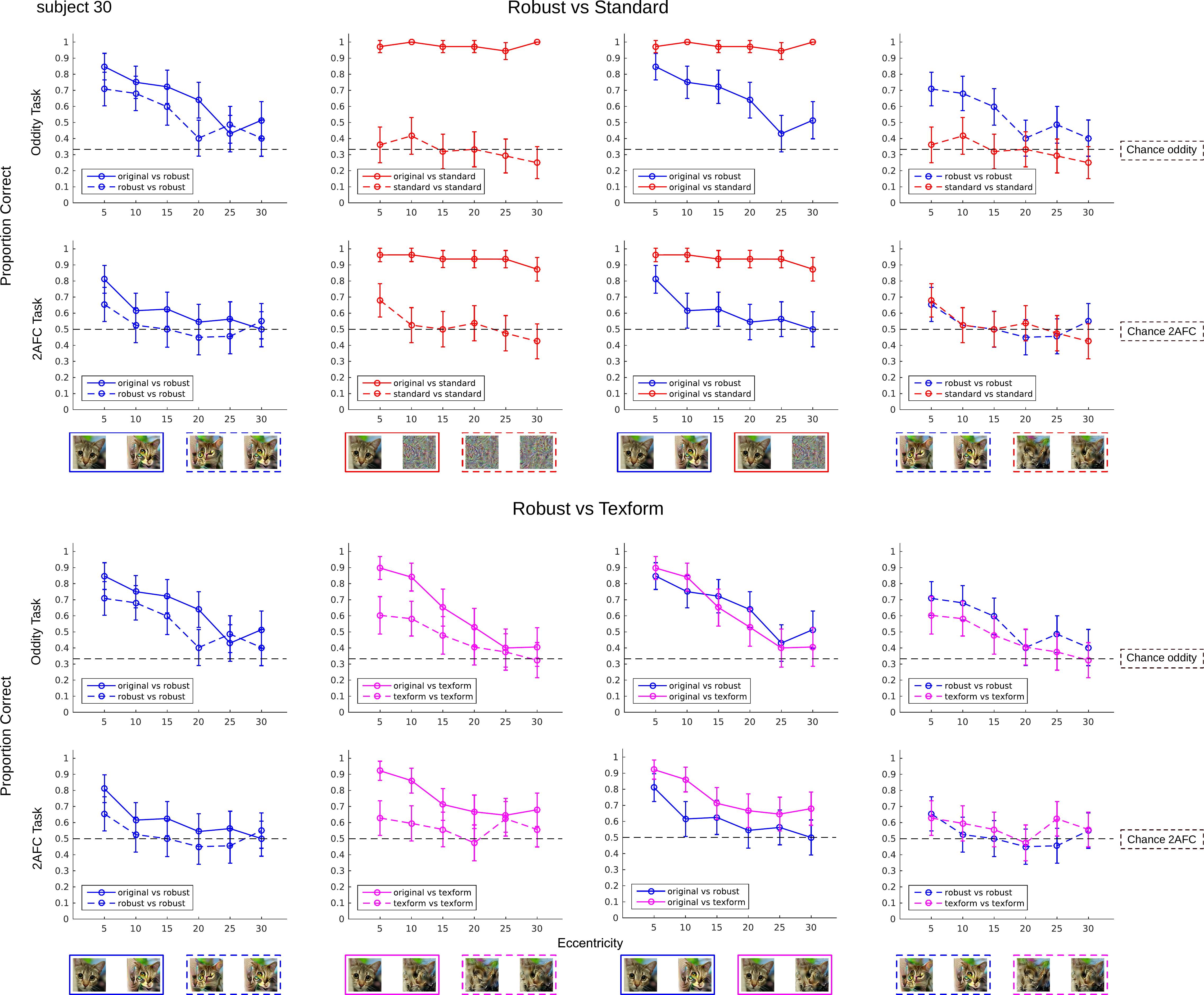}
    \caption{Subject 30}
    \label{fig:subject_30
    }
\end{figure}

\begin{figure}
    \centering
    \includegraphics[width=1.0\columnwidth]{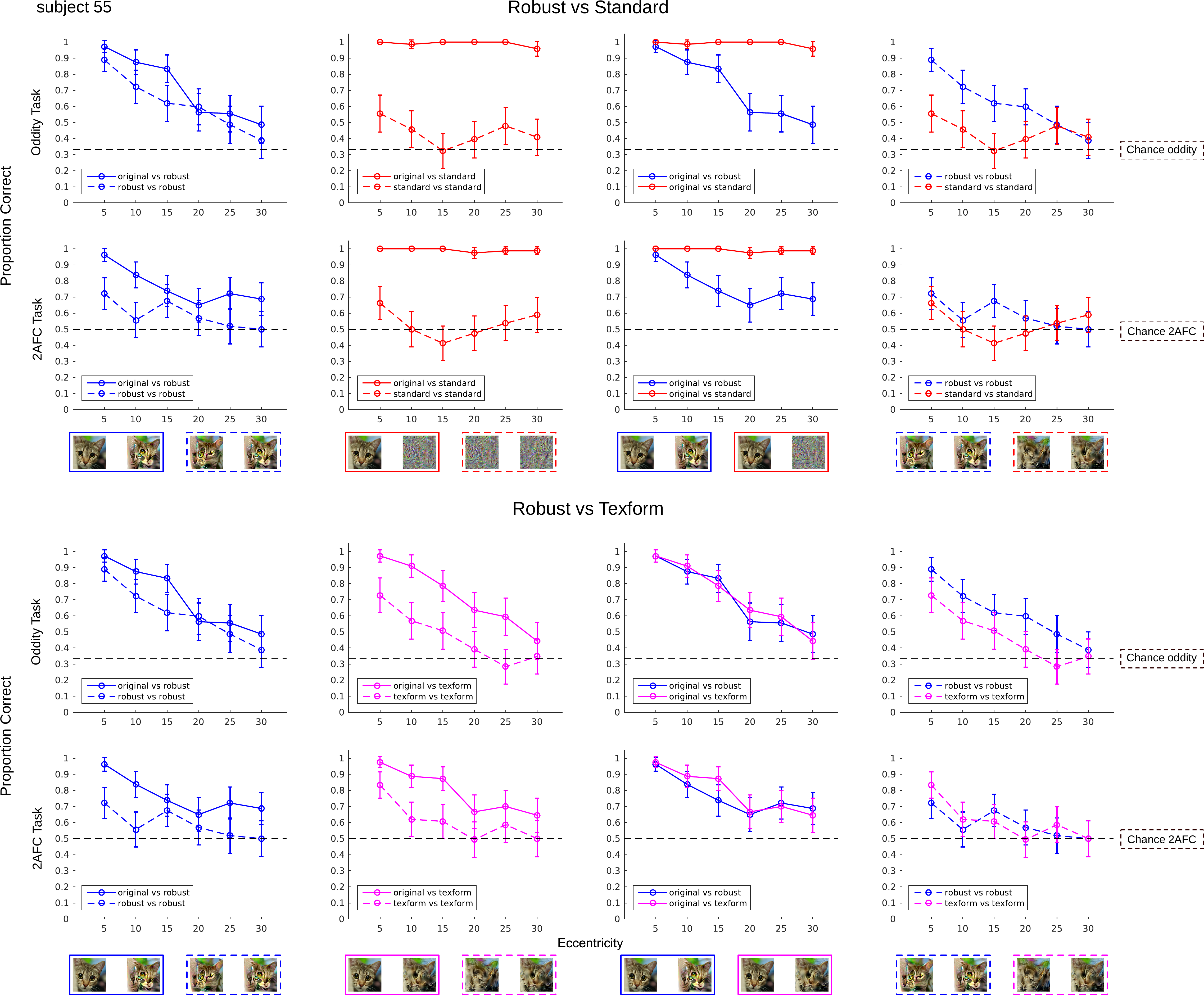}
    \caption{Subject 55}
    \label{fig:subject_55}
\end{figure}

\begin{figure}
    \centering
    \includegraphics[width=1.0\columnwidth]{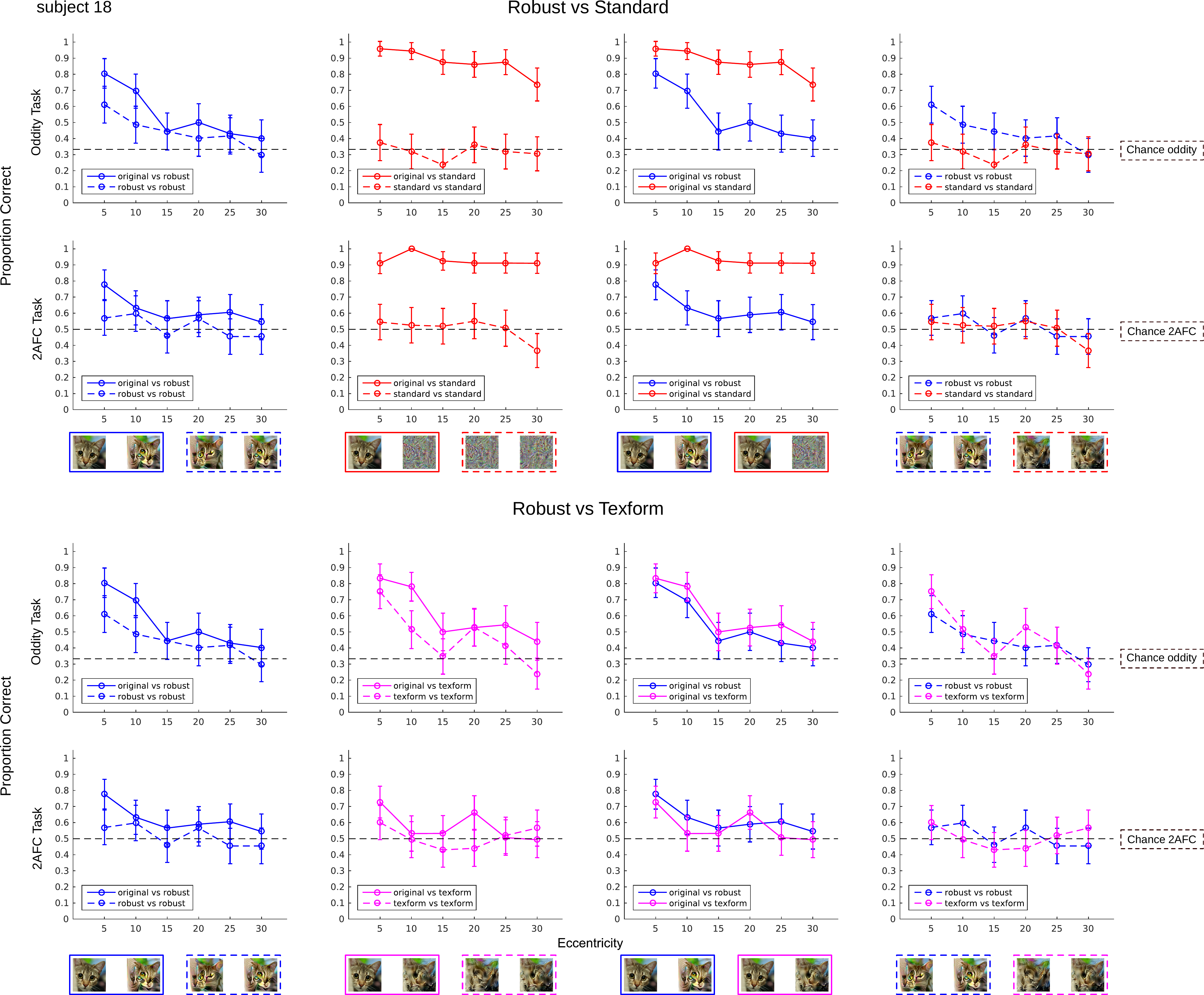}
    \caption{Subject 18}
    \label{fig:subject_18}
\end{figure}

\begin{figure}
    \centering
    \includegraphics[width=1.0\columnwidth]{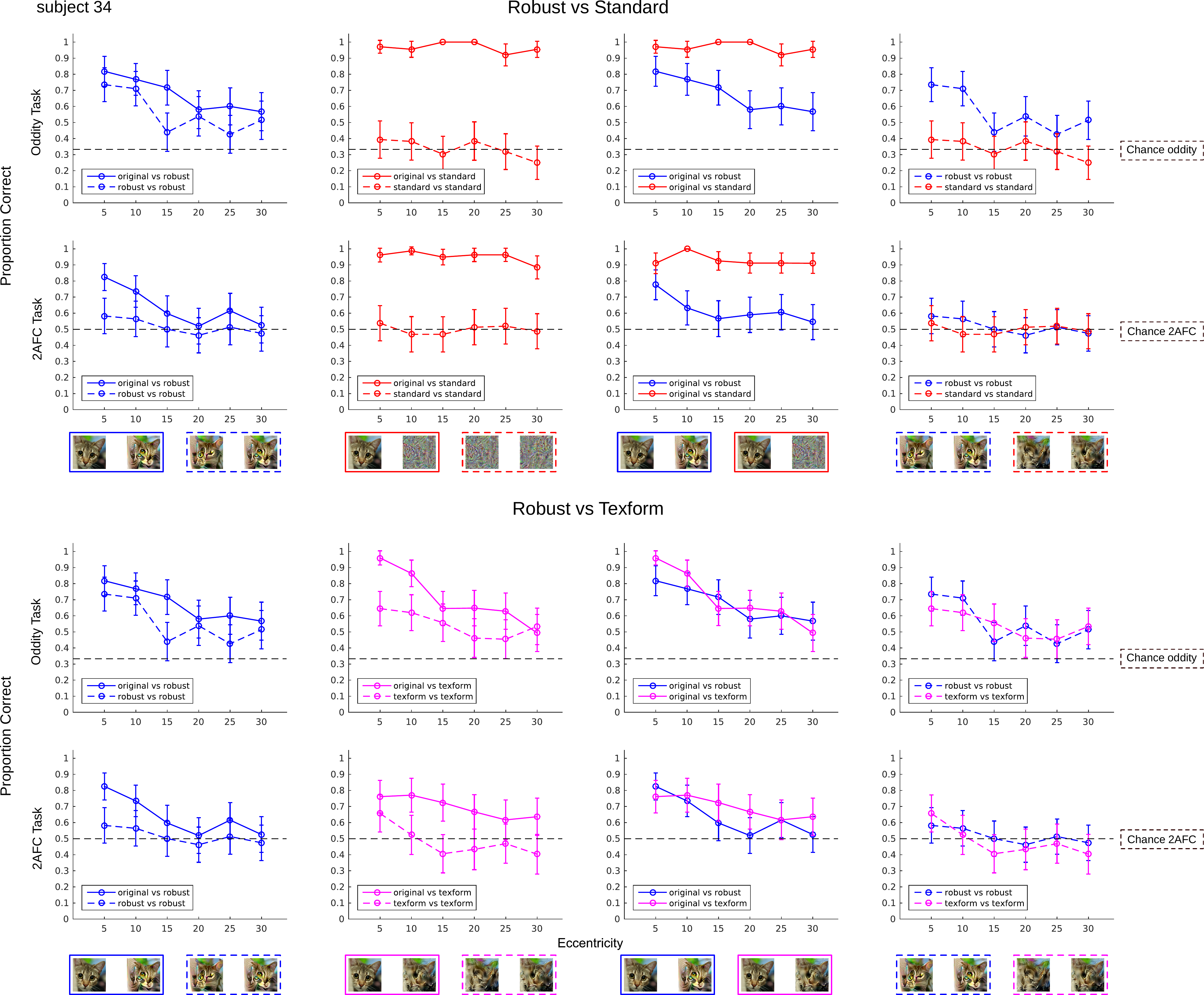}
    \caption{Subject 34}
    \label{fig:subject_34}
\end{figure}

\begin{figure}
    \centering
    \includegraphics[width=1.0\columnwidth]{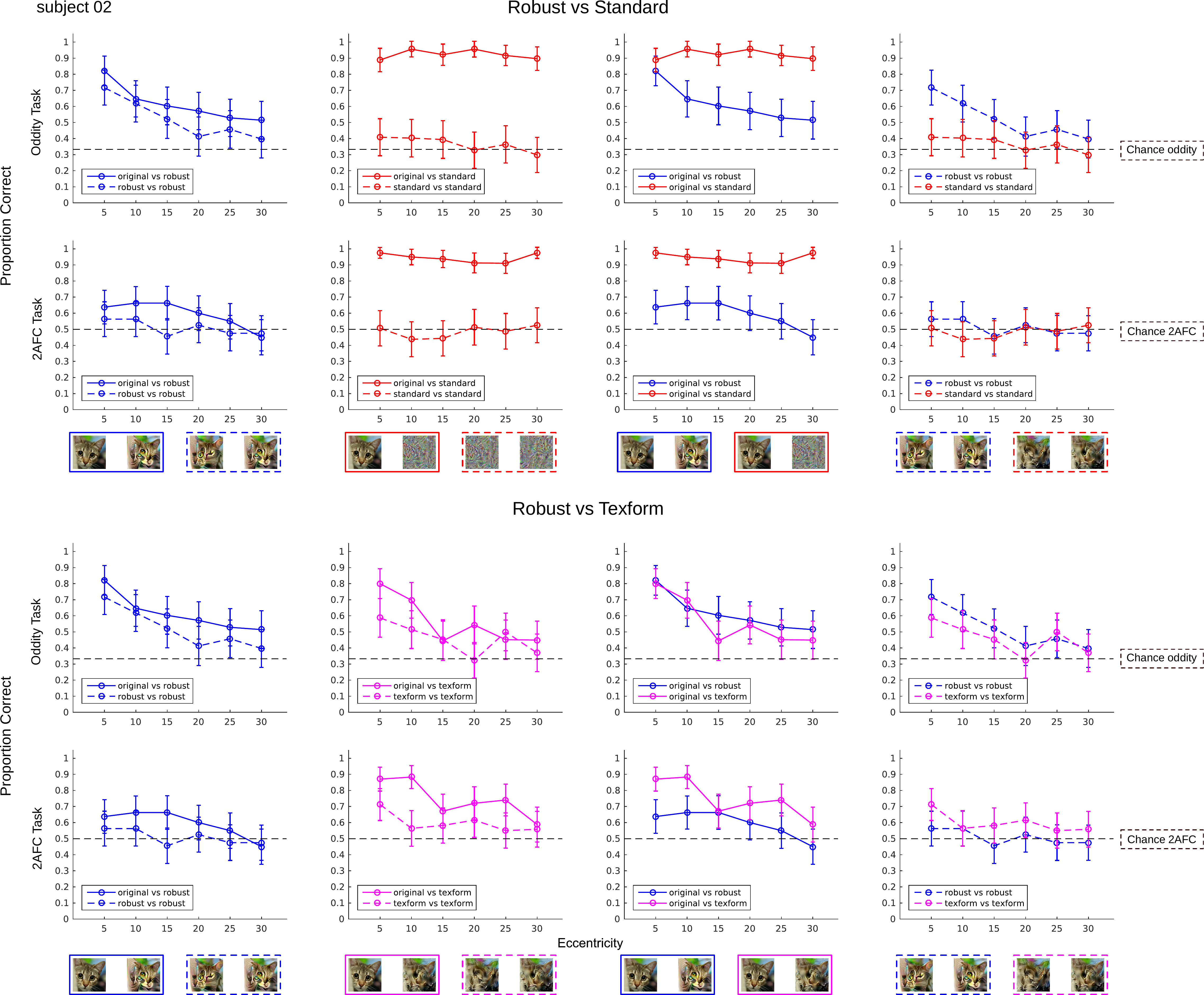}
    \caption{Subject 02}
    \label{fig:subject_02}
\end{figure}

\begin{figure}
    \centering
    \includegraphics[width=1.0\columnwidth]{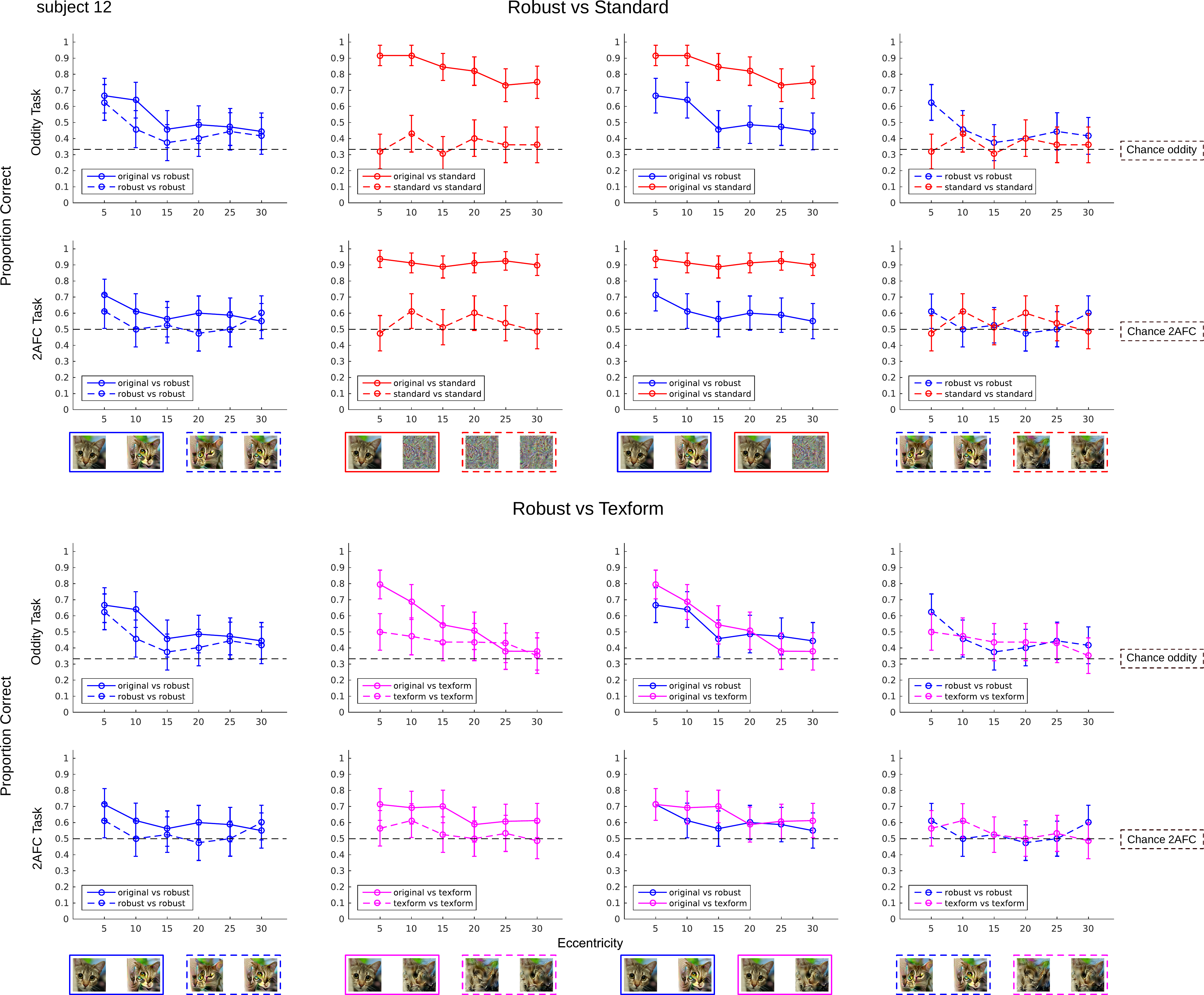}
    \caption{Subject 12}
    \label{fig:subject_12}
\end{figure}

\begin{figure}
    \centering
    \includegraphics[width=1.0\columnwidth]{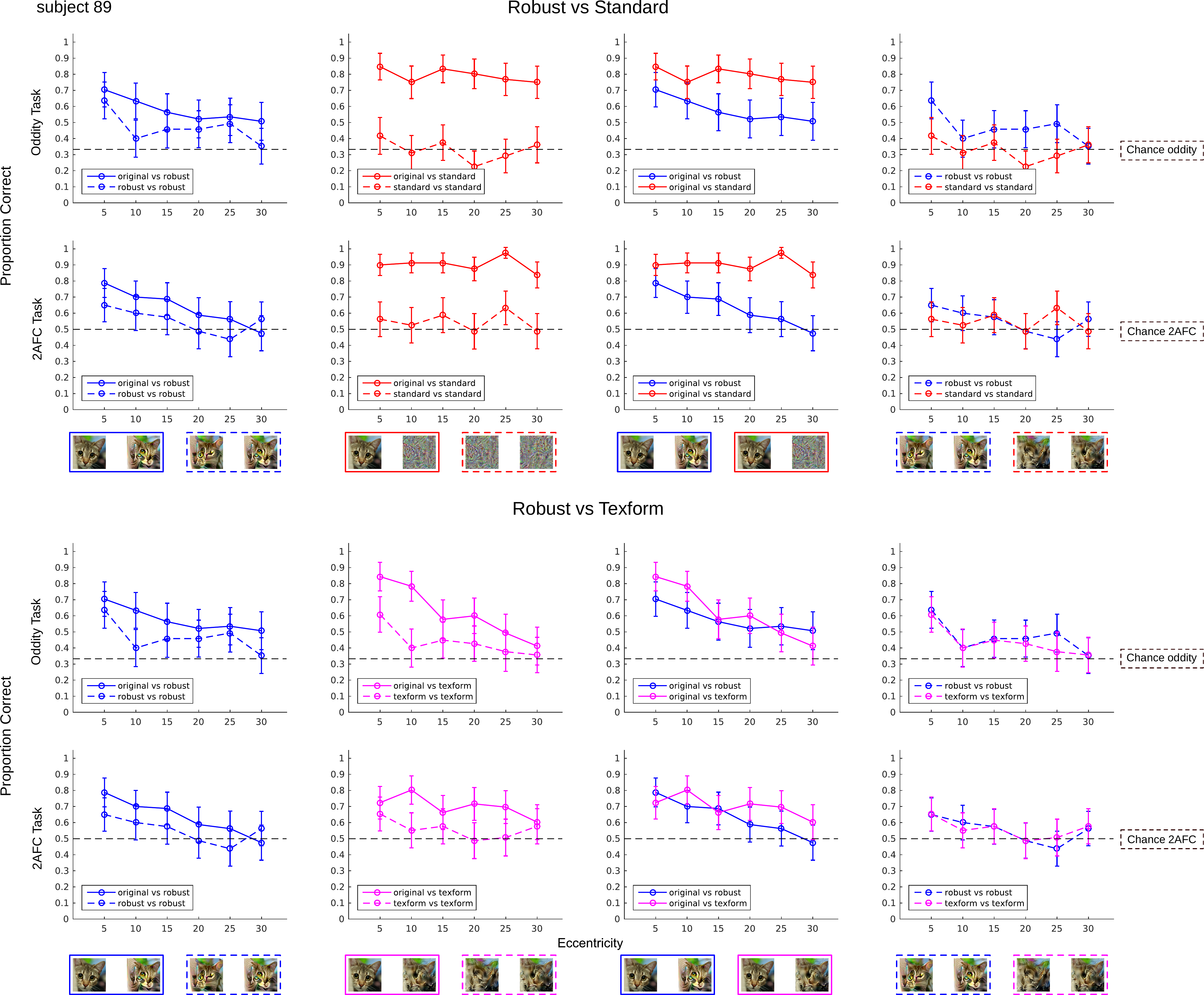}
    \caption{Subject 89}
    \label{fig:subject_89}
\end{figure}

\begin{figure}
    \centering
    \includegraphics[width=1.0\columnwidth]{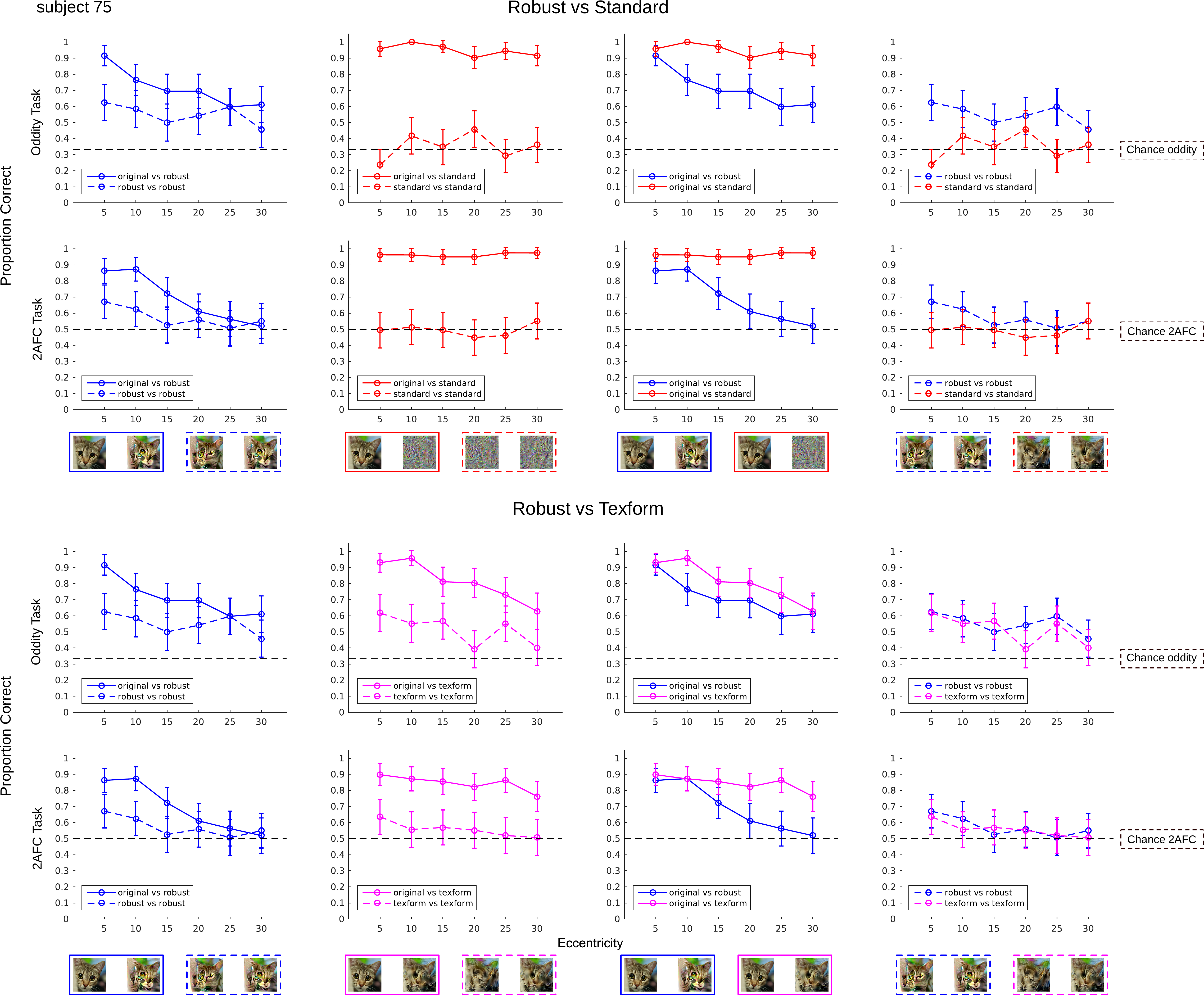}
    \caption{Subject 75}
    \label{fig:subject_75}
\end{figure}

\begin{figure}
    \centering
    \includegraphics[width=1.0\columnwidth]{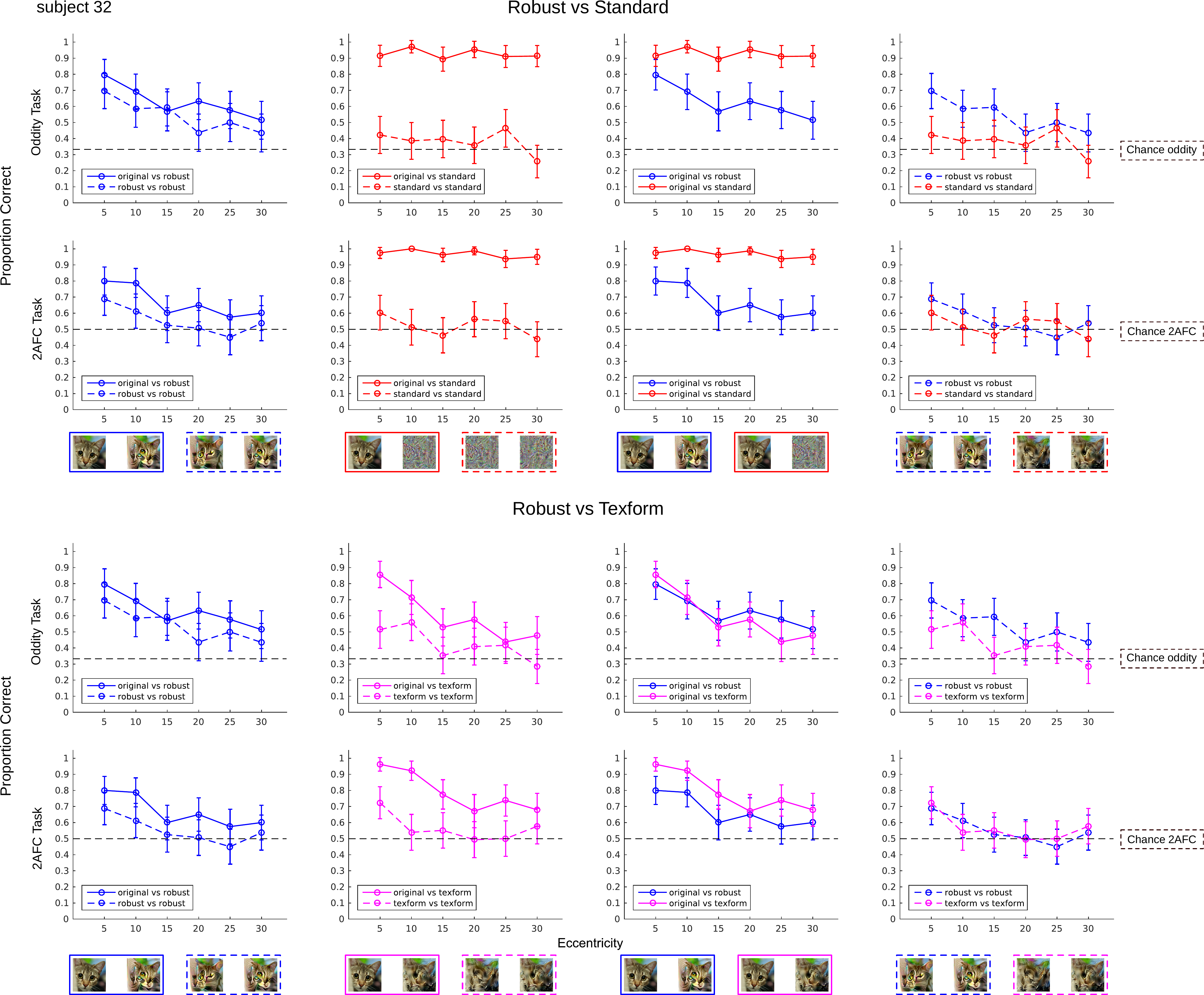}
    \caption{Subject 32}
    \label{fig:subject_32}
\end{figure}

\begin{figure}
    \centering
    \includegraphics[width=1.0\columnwidth]{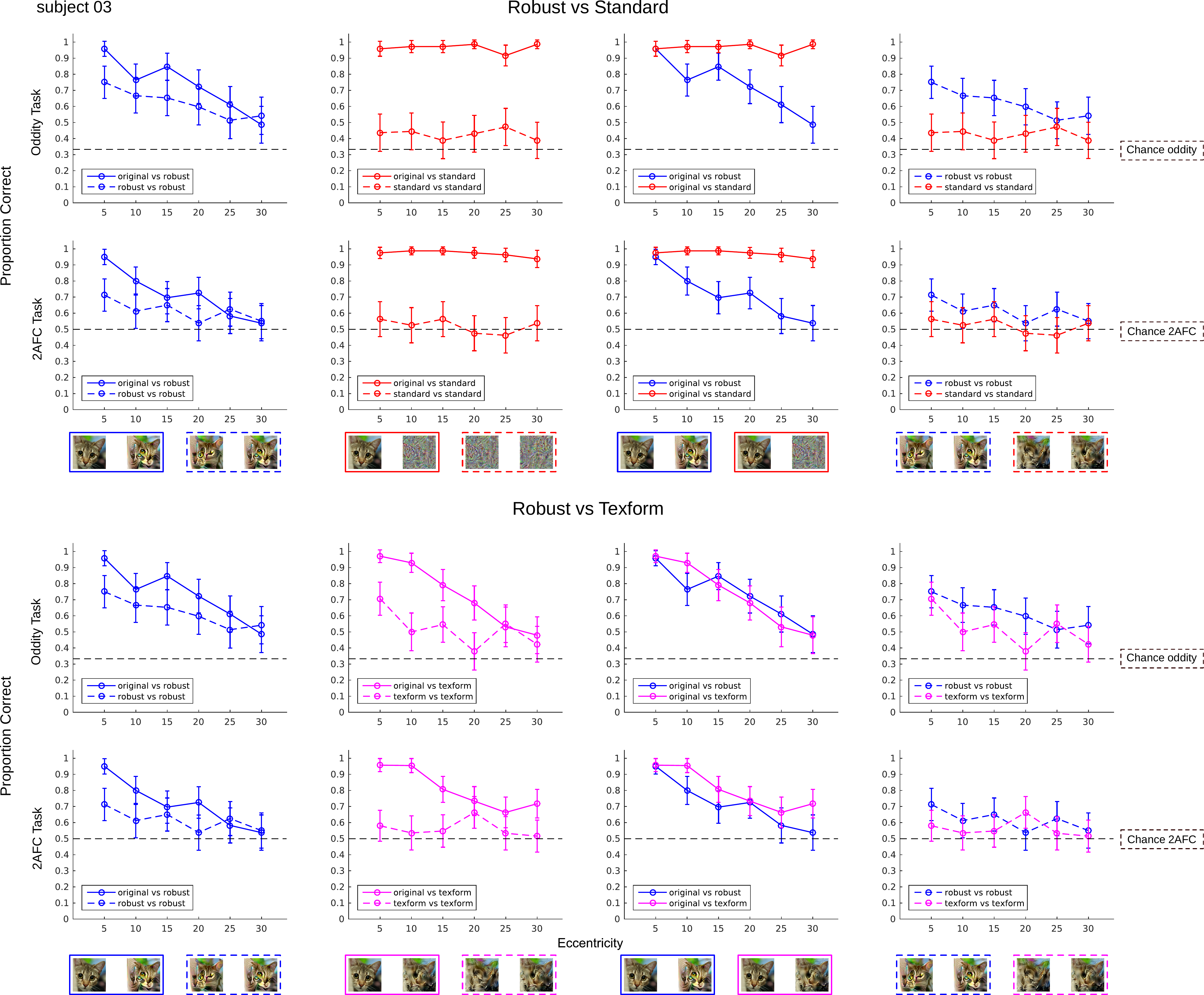}
    \caption{Subject 03}
    \label{fig:subject_03}
\end{figure}

\clearpage
\newpage

\end{document}